%% file: main.tex
\documentclass[10pt,twocolumn]{article}

\input{macros}

\title{FluxVLA Engine: A One-Stop VLA Engineering Platform for Embodied Intelligence}

\author{%
  \parbox{0.97\textwidth}{\centering
    \normalsize
    Yinhao~Li\textsuperscript{1},
    Weixin~Mao\textsuperscript{1},
    Zihan~Lan\textsuperscript{1},
    Jikun~Rong\textsuperscript{1},
    Qirui~Hu\textsuperscript{1},
    Yiming~Zhang\textsuperscript{2},
    \\[3pt]
    Weipeng~Deng\textsuperscript{3},
    Bowen~Shen\textsuperscript{1},
    Minzhao~Zhu\textsuperscript{1},
    Yiming~Mao\textsuperscript{1},
    Yan~Yang\textsuperscript{1},
    Chenguang~Cui\textsuperscript{4},
    \\[3pt]
    Hongyuan~Chen\textsuperscript{1},
    Xu~Huang\textsuperscript{1},
    Zheyi~Zhao\textsuperscript{1},
    Pinxi~Shen\textsuperscript{1},
    Bozhen~He\textsuperscript{1},
    Zhen~Fu\textsuperscript{1},
    \\[3pt]
    Yifan~Wang\textsuperscript{5},
    Zexin~Zhang\textsuperscript{6},
    Ang~Gao\textsuperscript{7},
    Haoyu~Chen\textsuperscript{1},
    Chengqi~Shi\textsuperscript{8},
    Wenda~Wu\textsuperscript{1},
    Hua~Chen\textsuperscript{1}\\[8pt]
    \small
    \textsuperscript{1}LimX Dynamics\quad
    \textsuperscript{2}Nankai University\quad
    \textsuperscript{3}The University of Hong Kong\\[2pt]
    \textsuperscript{4}University of Electronic Science and Technology of China\quad
    \textsuperscript{5}Beijing Institute of Technology\\[2pt]
    \textsuperscript{6}Xidian University\quad
    \textsuperscript{7}Zhejiang University\quad
    \textsuperscript{8}Xi'an Jiaotong University
  }%
}

\date{}

\begin{document}

\maketitle

\input{sections/00_abstract}
\input{sections/01_introduction}
\input{sections/02_related_work}
\input{sections/03_design_and_overview}

\input{sections/04_core_abstractions}
\input{sections/05_experiments}
\input{sections/06_conclusion}

\balance
\bibliographystyle{unsrt}
\bibliography{references}

\end{document}

%% file: macros.tex
\usepackage[T1]{fontenc}
\usepackage{microtype}
\usepackage{graphicx}
\usepackage{booktabs}
\usepackage{tabularx}
\usepackage{array}
\usepackage{amsmath,amssymb}
\usepackage{xcolor}
\usepackage{xspace}
\usepackage{enumitem}
\usepackage{etoolbox}
\usepackage{balance}
\usepackage{caption}
\usepackage[margin=0.72in,columnsep=0.24in]{geometry}
\usepackage{url}
\usepackage[colorlinks=true,allcolors=blue!55!black]{hyperref}
\usepackage[capitalize,noabbrev]{cleveref}

\setlist{nosep,leftmargin=*}

\newcommand{\fluxvla}{\textsc{FluxVLA}\xspace}

\newcolumntype{Y}{>{\raggedright\arraybackslash}X}
\newcolumntype{C}{>{\centering\arraybackslash}X}
\newcolumntype{P}[1]{>{\centering\arraybackslash}m{#1}}

\newcommand{\tablestyle}{%
  \centering
  \renewcommand{\arraystretch}{1.35}%
  \setlength{\extrarowheight}{2pt}%
}
\makeatletter
\renewcommand\section{\@startsection{section}{1}{\z@}%
  {-3.5ex \@plus -1ex \@minus -.2ex}%
  {2.3ex \@plus .2ex}%
  {\normalfont\Large\bfseries\raggedright}}
\renewcommand\subsection{\@startsection{subsection}{2}{\z@}%
  {-3.25ex \@plus -1ex \@minus -.2ex}%
  {1.5ex \@plus .2ex}%
  {\normalfont\large\bfseries\raggedright}}
\renewcommand\subsubsection{\@startsection{subsubsection}{3}{\z@}%
  {-3.25ex \@plus -1ex \@minus -.2ex}%
  {1.5ex \@plus .2ex}%
  {\normalfont\normalsize\bfseries\raggedright}}
\makeatother

\AtBeginEnvironment{thebibliography}{\raggedright}

%% file: sections/00_abstract.tex
\begin{abstract}
Vision--language--action (VLA) models, world--action models (WAMs), and
reinforcement learning (RL) methods are rapidly expanding the design space
of embodied policies, yet turning these algorithms into reliable robot systems
remains constrained by fragmented data formats, training stacks, evaluation
protocols, inference runtimes, and embodiment-specific interfaces. We present
\fluxvla Engine, an open, configuration-driven platform that turns
heterogeneous embodied-policy components into a reproducible
data-to-deployment workflow. Rather than introducing another policy model,
\fluxvla standardizes interfaces for datasets, vision--language model
(VLM) and world-model backbones, action heads, and robot operators. These
interfaces support distributed training, reward- or advantage-weighted
learning, simulation evaluation, and optimized inference. For online policy
improvement, an optional RLinf integration extends compatible pretrained
flow policies to online RL while preserving their native model and
preprocessing semantics. For data and interaction, the engine integrates
compositional dual-arm simulation, scalable automatic data generation, and
human-in-the-loop workflows for rollout, takeover, correction collection,
and reward annotation, decoupled from individual policy implementations.
For deployment, it combines Real-Time Chunking (RTC) with accelerated
inference backends, lightweight remote GPU serving, and configurable
trajectory post-processing. Together, these capabilities connect offline
learning, interaction-driven policy improvement, simulation validation,
and real-robot execution through shared, auditable interfaces. \fluxvla
therefore targets the engineering bottlenecks separating promising
embodied-learning algorithms from reproducible evaluation and dependable
deployment.
Code is available at \mbox{\url{https://github.com/FluxVLA/FluxVLA}}.
\end{abstract}

%% file: sections/01_introduction.tex
\section{Introduction}
\label{sec:introduction}

\begin{figure*}[!t]
  \centering
  \includegraphics[width=\textwidth]{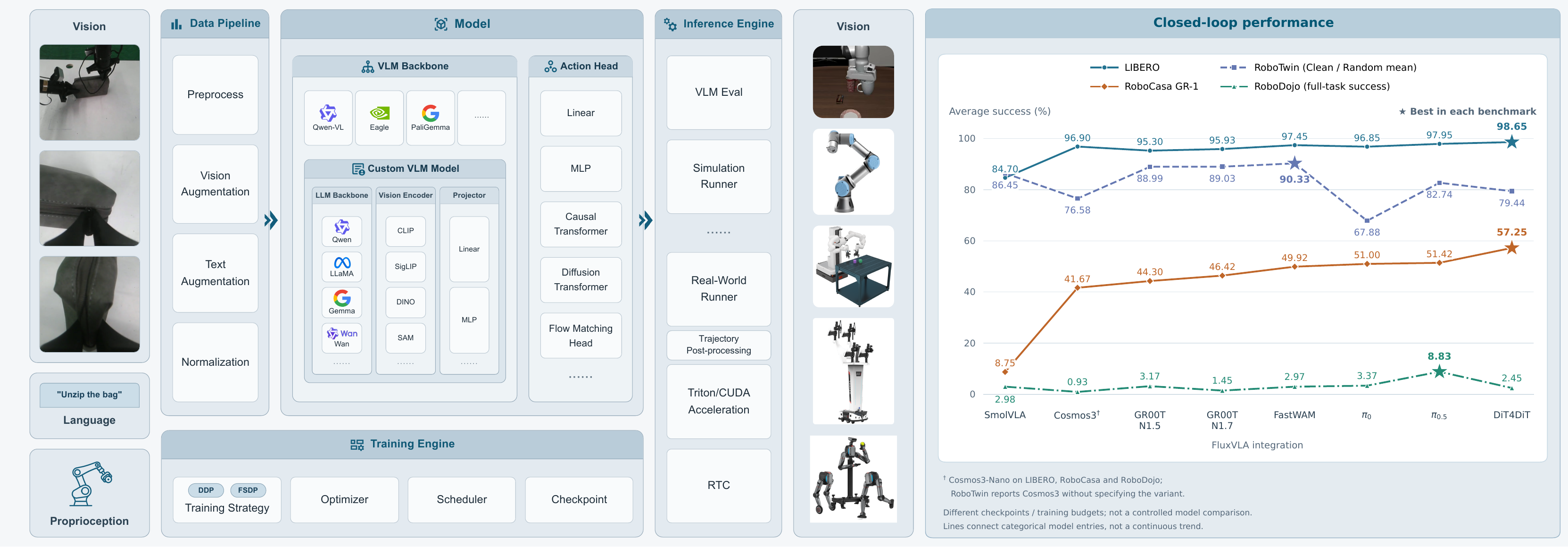}
  \caption{\fluxvla overview. Left: the modular data, model, training, and
  inference stack. Right: average closed-loop success rates (\%) across
  LIBERO, RoboCasa GR-1, RoboTwin, and RoboDojo. Stars mark the best reported
  result within each benchmark. Values are not normalized to a baseline;
  protocols and training budgets differ (\cref{sec:evaluation-protocol}).}
  \label{fig:system-overview}
\end{figure*}

\subsection{Motivation and problem setting}

Embodied policy learning now spans several complementary paradigms.
Vision--language--action (VLA) models map multimodal observations and language
instructions to robot actions~\cite{openvla2024,gr00t2025}; world--action
models (WAMs) additionally use predicted visual dynamics and temporally rich
world representations~\cite{dreamzero2026,dit4dit2026}; and offline
reinforcement learning (RL) improves policies from previously collected
experience without additional online data collection~\cite{offlinerl2020}.
Complementing fixed-data learning, online RL adapts pretrained policies
through new environment interactions~\cite{rlinfvla2025,pirl2025}. These
directions expand the range of tasks, data sources, and embodiments available
to robot learning, but an algorithm or checkpoint is not yet a robot system.
Turning a research implementation into repeatable physical behavior requires
consistent data representations, training and rollout infrastructure,
evaluation protocols, and robot-facing execution interfaces. These choices
are often embedded in project-specific code, making the path from a promising
policy to a working robot substantially harder than running the model itself.

The first gap appears before deployment. VLA, WAM, and RL implementations
frequently make different assumptions about dataset schemas and observation
formats (e.g., camera ordering and observation history), robot-state and
action representations (e.g., coordinate frames and prediction horizons),
and normalization statistics. Training launchers, checkpoint formats, and
evaluation runners are similarly coupled to specific models or benchmarks.
Integrating a new model or dataset therefore often requires rebuilding the
same adapters and execution logic. Results obtained through different
preprocessing pipelines, evaluator versions, or rollout protocols can also
be difficult to compare. This fragmentation slows iteration and reproduction
and obscures whether an observed gain comes from the learning algorithm or
from differences in the surrounding system.

Physical deployment introduces a second, stricter gap. Success in simulation
does not guarantee reliable real-robot operation: a deployment stack must
reconcile model outputs with embodiment-specific control modes and hardware
interfaces while handling observation staleness, inference and communication
latency, and motion constraints. This is especially challenging for
action-chunking VLA policies: inference may be slower than the robot control
loop, and executing one chunk asynchronously while predicting the next can
produce discontinuities at chunk boundaries. Even a temporally consistent
prediction may violate embodiment-specific motion constraints when executed
directly. Hosting a large policy on a remote GPU relaxes robot-side compute
requirements, but adds serialization, transport, timeout, and recovery
concerns. Runtime acceleration, temporal coordination, trajectory
post-processing, serving, and robot operators must therefore be treated as
first-class parts of the learning system rather than deployment details
added after training.

Bridging these gaps requires a closed engineering loop spanning data,
training, simulation, and real-robot deployment, with failures and human
corrections fed back into training. \fluxvla is designed around this loop.
Rather than proposing another policy architecture, it provides a common
engineering layer for composing models, data, learning workflows, evaluators,
and robot interfaces into reproducible systems.
\Cref{fig:system-overview} summarizes the platform architecture and
representative simulation results. Visual, language, and proprioceptive
inputs pass through a configurable data pipeline to integrated or custom
vision--language model (VLM) backbones and replaceable action heads. The
training engine manages distributed optimization, scheduling, and
checkpointing; the inference engine supports evaluation and physical-robot
runners with accelerated kernels, Real-Time Chunking (RTC), and trajectory
post-processing across embodiments.

The same policy and preprocessing interfaces also support online policy
improvement through an optional RLinf frontend~\cite{yu2026rlinf,rlinfvla2025}.
\fluxvla retains native model construction and observation--action transforms,
while RLinf manages rollout collection and policy optimization. In the paired
SmolVLA evaluations reported in \cref{tab:rl-post-training-results}, task-wise
PPO post-training increases mean success from 66.2\% to 76.2\% on LIBERO-10
and from 45.9\% to 61.3\% on ten RoboTwin Clean tasks after 40 outer steps.
These gains of 10.0 and 15.4 percentage points demonstrate an online
improvement path under matched SFT--RL evaluation conditions, not a new RL
algorithm or a controlled comparison across policy families.

\subsection{Scope}

\fluxvla is a configuration-driven engineering platform spanning data
preparation to physical-robot deployment. Its scope is the reusable
infrastructure and interfaces connecting these stages. The platform
integrates existing VLA and WAM architectures, offline and reward-weighted
learning, optional online RL post-training, benchmarks, simulators, reward
models, and robot SDKs; it does not claim their underlying algorithms as new.
We evaluate the platform through integration coverage, execution consistency,
reproducibility, runtime efficiency, and demonstrated deployment workflows.
Closed-loop success rates in \cref{fig:system-overview} provide evidence of
integrated-policy performance, but do not by themselves establish execution
correctness. They reflect each checkpoint and its training and evaluation
protocol, so they should not be interpreted as a controlled architecture
ranking. Paired SFT--RL evaluations are reported separately under matched
conditions, and implementation-level checks assess inference parity and
rollout--learner consistency (\cref{sec:rl-results,sec:runtime-results}).

\subsection{Contributions}

We present \fluxvla to bridge the gap between rapidly evolving embodied-policy
models and the fragmented engineering stacks needed to train, evaluate, and
deploy them. The platform provides a reusable full-stack foundation organized
around four engineering principles: unification, modularity, deployability,
and reproducibility. Our main contributions are:

\begin{enumerate}
\item \textbf{A unified, configuration-driven embodied-policy workflow.}
Configuration-defined interfaces connect data processing, model construction,
supervised learning, evaluation, and deployment. An optional RLinf frontend
composes native model and preprocessing specifications with RL rollout and
optimization configurations, extending the workflow to online policy
improvement.

\item \textbf{A modular and extensible system architecture.}
Registry-based interfaces decouple major data, model, execution, and robot
components, enabling compatible methods to reuse the shared workflow while
localizing model- and embodiment-specific adaptation.

\item \textbf{A deployment-ready and optimized execution stack.}
A unified runtime serves simulation and physical robots through local or
remote inference, with GPU acceleration, RTC, and trajectory post-processing
for efficient and temporally coordinated control across embodiments.

\item \textbf{A reproducible experimentation pipeline.}
Resolved configurations, checkpoint and preprocessing artifacts, resumable
training, and structured rollout records support reproducible evaluation.
Matched protocols and available seed controls enable paired measurements
without assuming identical determinism across simulators.
\end{enumerate}

%% file: sections/02_related_work.tex
\section{Related Work}
\label{sec:related}

\fluxvla lies at the intersection of vision--language--action (VLA) and
world--action modeling (WAM), offline and online policy improvement,
robot-learning infrastructure, and deployment systems. These areas are tightly
coupled in a working robot stack but address different questions. Model papers
study how observations and instructions are mapped to actions or future visual
dynamics; offline learning studies how logged experience can improve a policy;
datasets and benchmarks determine what experience and evaluation protocols are
available; and engineering platforms determine whether these methods can be
trained, compared, and deployed through one reproducible path. We review these
lines separately and then delimit the claims of this report.

\subsection{Engineering platforms for model development}

Mature perception toolboxes established an important precedent for treating
software architecture as a systems contribution.
MMDetection~\cite{mmdetection2019} organizes detectors from many papers through registries,
reusable components, declarative configurations, and common training and
evaluation tools. MMDetection3D~\cite{mmdetection3d2020} extends the pattern to
heterogeneous 3D detectors, datasets, and sensor modalities. Their
relevance here is methodological: support breadth becomes scientifically useful
when alternative components obey explicit contracts and can be evaluated under
controlled configurations, rather than being isolated reproductions with
unrelated scripts.

Robot-learning libraries bring this toolbox idea closer to embodied systems.
robomimic provides standardized implementations and empirical comparisons for
learning manipulation policies from offline human demonstrations
~\cite{robomimic2021}. LeRobot~\cite{lerobot2026} spans hardware interfaces,
data collection and storage, policy implementations, training, visualization,
and asynchronous inference. StarVLA targets Lego-like VLA development through
modular model abstractions, reusable training strategies, and unified benchmark
interfaces~\cite{starvla2026}. These systems
substantially lower the cost of reproduction, yet a shared model API alone does
not guarantee consistency across preprocessing, checkpoint restoration,
evaluation, serving, and physical execution.

\fluxvla follows this toolbox tradition while making the complete
configuration-resolved execution path its unit of integration. A supported
method therefore includes its dataset and ordered transforms, model
construction, distributed runner, self-contained checkpoint, evaluator or
inference runner, and simulator or robot operator. The report evaluates whether
heterogeneous policy families can share stable interfaces from offline data to
real-device execution, rather than treating the number of registered model
names as the primary evidence of support.

\subsection{Embodied policy architectures}

Early generalist policies demonstrated that transformer sequence models can
represent many language-conditioned robot skills in a single network. RT-1
tokenizes observations and actions for scalable real-world control
~\cite{rt12022}, while RT-2 co-fine-tunes vision--language models on web and
robot data so that semantic knowledge can influence action prediction
~\cite{rt22023}. Subsequent open-source systems expose several distinct action
interfaces that an engineering platform must preserve rather than flatten into
one nominal ``VLA'' abstraction.

\paragraph{Discrete and autoregressive actions.}
OpenVLA casts continuous control as next-token prediction by discretizing
actions and adapting a pretrained vision--language model on robot trajectories
~\cite{openvla2024}. FAST combines a DCT-based frequency representation,
quantization, and byte-pair encoding to compress action chunks into
substantially fewer tokens than per-timestep discretization, making
autoregressive learning on high-frequency control data more
tractable~\cite{fast2025}. Such methods inherit language-model
training and decoding machinery but introduce policy-specific choices about
quantization, token ordering, de-tokenization, and sequential inference. They
motivate explicit tokenizer and action-space contracts instead of conversions
hidden inside a particular backbone.

\paragraph{Continuous action chunks and generative policies.}
Action Chunking with Transformers predicts temporally extended action
sequences and temporally ensembles overlapping predictions~\cite{act2023}.
Diffusion Policy instead represents multimodal visuomotor behavior with a
conditional denoising process~\cite{diffusionpolicy2023}. Octo demonstrates an
open-source generalist transformer policy trained across heterogeneous robot
datasets~\cite{octo2024}. More recent VLAs combine large pretrained
representations with continuous generative action heads: $\pi_0$ couples a
vision--language model to a flow-matching action expert~\cite{pi02024},
$\pi_{0.5}$ extends the approach toward heterogeneous data and open-world
generalization~\cite{pi052025}, and GR00T N1 combines a vision--language
reasoning module with a diffusion-transformer action module for generalist
humanoid control~\cite{gr00t2025}. SmolVLA investigates a related interface
under stronger efficiency and accessibility constraints~\cite{smolvla2025}.
LLaVA connects a pretrained visual encoder to a language model through a
learned projection layer, exemplifying modular visual--language integration
~\cite{llava2023}; action generation remains policy-specific.

These policies differ not only in backbone size. They require distinct action
horizons, noise or time parameterizations, state-conditioning paths, attention
masks, normalization conventions, and inference loops. Consequently,
interchangeability should be claimed only at an interface whose tensors and
semantics actually agree; a common base class is not evidence that every
backbone and action head can be freely combined.

\paragraph{World--action models.}
WAMs use video or latent dynamics during policy learning and may jointly model
future visual trajectories and actions. DreamZero adapts pretrained video
generation into a zero-shot policy that produces future visual and action
trajectories~\cite{dreamzero2026}. DiT4DiT jointly models video dynamics and
actions with diffusion transformers~\cite{dit4dit2026}. Fast-WAM separates the
benefit of video co-training from test-time imagination and omits future-video
generation during action inference~\cite{fastwam2026}. Relative to conventional
VLA heads, these methods introduce video tokenization, temporal packing,
cross-modal attention, additional training targets, and substantially different
memory and scheduling behavior. Their coexistence with autoregressive and
flow-based VLAs motivates \fluxvla's separation of backbones, temporal or world
modules, action heads, losses, and inference procedures behind typed data
contracts.

\subsection{Offline and online policy improvement}

Offline reinforcement learning seeks to improve a policy from a fixed dataset
without additional online data collection~\cite{offlinerl2020}. Its central
difficulty is distribution shift: value estimates can become unreliable for
actions insufficiently represented by the behavior data. Conservative
Q-Learning regularizes learned values against this failure mode
~\cite{cql2020}, whereas Implicit Q-Learning avoids explicitly evaluating
out-of-distribution actions and extracts a policy through advantage-weighted
regression~\cite{iql2022}. In an adjacent interactive setting, DAgger
aggregates expert action labels for states visited by the learned policy, directly
addressing the train--execution state-distribution mismatch
~\cite{dagger2011}.

For large robot policies, reward models, progress estimates, and sample
reweighting provide practical bridges between supervised imitation and policy
improvement. ARM, for example, uses learned advantage estimates to reweight
behavior-cloning updates for long-horizon manipulation~\cite{arm2026}. These
methods require more than replacing a loss function: datasets must retain
reward, progress, source-policy, and intervention metadata; transforms and
collators must preserve sample weights; runners must coordinate policy and
reward-model computation; and evaluation must distinguish algorithmic
improvement from changes in the underlying data mixture. \fluxvla therefore
treats offline and reward-guided learning as extensions of the same data,
model, and runner contracts used for supervised VLA and WAM training.

Online RL instead improves a policy using newly collected interactions.
RLinf provides flexible, large-scale RL execution through macro-to-micro
flow transformation~\cite{yu2026rlinf}.
RLinf-VLA provides shared infrastructure for rollout generation, distributed
optimization, and simulator integration~\cite{rlinfvla2025}. For flow-based
policies, $\pi_{\mathrm{RL}}$ studies stochastic denoising formulations that
make exploration and likelihood-based updates tractable~\cite{pirl2025}.
\fluxvla builds on this line of work through an optional RLinf frontend for
its native policies, rather than introducing a new RL algorithm. This
frontend reuses model construction and observation--action transforms while
delegating rollout orchestration and policy optimization to the RL backend.

\subsection{Robot data and evaluation benchmarks}

Large heterogeneous datasets make cross-embodiment learning possible while
turning data semantics into a first-order systems concern. RLDS introduced a
common episodic representation for sharing sequential decision data
~\cite{rlds2021}; Open X-Embodiment aggregates trajectories from multiple
institutions and robot platforms and uses them to train RT-X models
~\cite{openx2023}; and LeRobot~\cite{lerobot2026} provides a practical
representation and tooling stack for storing, visualizing, streaming, and
processing robot episodes. A common container improves portability but does not by
itself reconcile camera order, control frequency, coordinate frames,
proprioceptive state, action parameterization, normalization statistics,
episode boundaries, or language annotations. These conversions must remain
explicit, ordered, and versionable if training and inference are to implement
the same observation--action contract.

Simulation benchmarks enable repeatable closed-loop testing across different
task distributions and capabilities. LIBERO organizes language-guided
manipulation into suites for knowledge transfer and lifelong learning
~\cite{libero2023}. RoboCasa provides large-scale household simulation,
diverse kitchen scenes and assets, and everyday manipulation
tasks~\cite{robocasa2024}; our evaluation uses the RoboCasa GR-1 Tabletop
task suite built on this framework~\cite{gr00t2025}.
RoboTwin 2.0 complements these environments with bimanual manipulation tasks,
automated data generation, and structured domain
randomization~\cite{robotwin2025}. RoboDojo's simulation benchmark further
assesses generalization, memory, precision, long-horizon execution, and
open-vocabulary instruction following~\cite{robodojo2026}.

Success rates should be interpreted within each benchmark's task distribution
and protocol rather than directly compared across benchmarks. Controlled
comparisons require consistent evaluation conditions and checkpoint-selection
rules, with rollout budgets and statistical uncertainty reported explicitly.
These requirements motivate \fluxvla's explicit data transformations and
benchmark-specific evaluation protocols. The evaluated task sets, result
provenance, and available protocol details are presented in
\cref{sec:evaluation-protocol,sec:simulation-results}.

\subsection{Inference and physical deployment}

Physical execution changes the objective from merely predicting an action to
delivering safe and temporally consistent commands at a stable control rate.
Although action chunking amortizes policy queries, a robot may continue
executing an outdated chunk while the next prediction is being computed.
The original Real-Time Chunking method addresses this asynchronous setting
with inference-time guidance-based inpainting, using a vector--Jacobian product
to condition each newly generated flow-policy chunk on actions already
committed for execution~\cite{rtc2025}. Training-Time Action Conditioning later
moves this capability into the learned policy: training simulates inference
delay so that inference can directly condition on the known action
prefix~\cite{trainingtimertc2025}. Both routes enable asynchronous policy
inference while reducing discontinuities at chunk boundaries.
LeRobot~\cite{lerobot2026} likewise identifies asynchronous inference as part
of an end-to-end learning stack.

Prediction-time consistency does not by itself guarantee an executable motion.
After denormalization, a chunk may still violate execution-side motion
requirements or connect poorly to the trajectory already in progress.
Trajectory post-processing addresses this boundary after prediction. It
complements RTC: RTC shapes the next prediction, while post-processing shapes
the command ultimately sent to the robot.

This temporal coordination must be combined with systems-level inference
optimization. Reduced-precision execution, CUDA Graph capture, fused operators,
and optimized attention kernels can increase model throughput. Realtime-VLA
demonstrates a complementary systems path by running a $\pi_0$-level multi-view
VLA at 30 Hz on a single consumer GPU and proposes a full streaming inference
framework for higher-frequency control~\cite{realtimevla2025}. Local and
remote inference runners then determine how predictions are delivered to the robot.
Remote serving additionally introduces serialization, network latency,
timeouts, health checks, reset semantics, and recovery behavior. The deployment
interface must therefore track observation timestamps, committed actions,
execution horizons, communication latency, and failure handling. In this
report, model-only throughput is consequently separated from end-to-end control
frequency. Inference acceleration, RTC, and trajectory post-processing are
therefore treated as complementary components of the deployment stack.

\subsection{Positioning and claim boundary}

This report studies \fluxvla as an engineering platform. The platform owns the
configuration and build path, unified data adapters and composable transforms,
component registries, training and evaluation runners, checkpoint and serving
plumbing, runtime optimizations, and robot-facing operators implemented in the
released code. It integrates---but does not claim authorship of---upstream
architectures, pretrained weights, datasets, simulators, benchmarks, robot
SDKs, or their algorithmic contributions.

Four boundaries follow. First, the report does not propose a universal VLA or
WAM architecture; it studies whether heterogeneous architectures can coexist
behind stable contracts. Second, it does not introduce a benchmark and follows
the protocols and limitations of the suites it integrates. Third, a registry
entry is not evidence of end-to-end support: training requires a reproducible
checkpoint, evaluation requires a fixed protocol and raw rollouts, and
deployment requires a runnable operator and measured execution. Finally,
performance claims are confined to matched data, checkpoints, evaluators,
hardware, and trial counts. Measurements from different settings are reported
as separate system results rather than evidence of universal superiority.

%% file: sections/03_design_and_overview.tex
\section{System Design and Overview}
\label{sec:overview}

A VLA engineering system must accommodate two forms of change at the same
time. The learning stack evolves as new visual--language backbones, action
representations, and optimization strategies are introduced, while the
execution stack varies across datasets, simulators, inference hardware, and
physical robots. A design that couples these choices quickly degenerates into
parallel model-specific code paths: each integration obtains its own data
loader, launcher, evaluator, and deployment script, and improvements made in
one path do not propagate to the others. \fluxvla is designed to avoid this
fragmentation by separating component selection from component implementation
and by expressing the complete experiment lifecycle through common
configuration and execution contracts.

This chapter presents the system from the outside in. We first describe the
overall design, including the registry and configuration mechanisms, the
separation of modules, the train--evaluate lifecycle, and the construction of
self-contained inference artifacts. The subsequent sections follow the path of
an experiment: data preparation and transformation, model composition,
simulation evaluation, real-robot inference, inference acceleration,
Real-Time Chunking (RTC), and trajectory post-processing. We then describe
online RL post-training, which feeds newly collected rollout experience back
into policy optimization. This ordering makes explicit how
the same configuration and intermediate representations connect offline
training to closed-loop execution, while keeping benchmark- and
robot-specific logic at the boundaries of the system. The complete workflow is
summarized in \cref{fig:system-overview}.

\subsection{Overall design}
\label{sec:overall-design}

\paragraph{Design objective.}
The primary objective of \fluxvla is to make a VLA experiment a composable and
reconstructable system rather than a collection of scripts. The framework is
therefore organized around four principles. \textbf{Unification} places data,
model, optimization, evaluation, and deployment choices in one experiment
description. \textbf{Modularity} separates implementations behind explicit
construction and runtime interfaces. \textbf{Deployability} preserves the model
and preprocessing path while replacing only the environment- or robot-facing
execution layer. \textbf{Reproducibility} records the resolved experiment state
and exports the artifacts required to evaluate or deploy the trained policy.
These principles are implemented jointly: a registry without a complete
configuration does not reproduce an experiment, while a configuration without
stable module boundaries merely moves hard-coded coupling into a different
file.

\paragraph{Registry-based construction.}
\fluxvla uses type-keyed registries to decouple component selection from
implementation, providing the construction boundary between an experiment
configuration and its components. Dedicated registries cover tokenizers,
transforms, datasets, collators, vision and language backbones, VLM backbones,
projectors, action heads, top-level VLA models, runners, metrics, processors,
robot operators, optimizers, schedulers, evaluators, and weighting modules. A
component registers a symbolic name, and a category-specific builder resolves
the \texttt{type} field in a configuration block to the corresponding Python
class or factory. The remaining fields are passed as constructor arguments.
Nested components are built at their owning boundary: for example, a VLA
constructs its registered backbone, projector, and action head, while a runner
constructs the VLA, collator, metric, and optional evaluator.

This mechanism removes model selection from the training and inference entry
points. The launcher asks for a dataset or runner to be built; it does not
contain a branch for every supported policy family. Adding an implementation
therefore consists of registering the new component and supplying a
configuration that satisfies the existing contract, rather than copying the
entire workflow. Importantly, registration establishes a construction
interface, not unrestricted interchangeability. Components must still agree on
field names, tensor shapes, action dimensions, temporal horizons, and lifecycle
semantics. Model-specific adapters remain appropriate when those semantics
differ, but the differences are localized rather than propagated through the
whole stack.

\paragraph{Modular system boundaries.}
The platform separates five major responsibilities. The data layer maps stored
episodes and online observations into a canonical sample dictionary through
composable transforms and collators. The model layer maps that dictionary to
losses during training or action chunks during inference. The execution layer
owns distributed setup, optimization, checkpointing, evaluation loops, and
runtime state. The serving layer transports observations and actions when the
policy and robot execute in different processes or machines. The execution
layer can additionally apply trajectory post-processing after denormalization,
keeping motion constraints and cross-chunk stitching independent of the policy
architecture. Finally, the operator layer translates between the
framework-level state/action contract
and a simulator or robot SDK. This decomposition allows, for example, a new
robot operator to reuse an existing policy and inference runner, or a new
action head to reuse the same dataset and distributed training engine.

The boundary between runners and operators is particularly important. Runners
own model-side decisions such as observation assembly, action-chunk scheduling,
local versus remote prediction, and RTC state. Operators own hardware-facing
I/O, including sensor acquisition and command transport. Keeping these
responsibilities separate prevents robot communication code from becoming a
hidden dependency of the model and lets simulation and physical execution
share the same policy-facing interface.

\paragraph{Configuration as an executable experiment specification.}
The native \fluxvla workflow uses hierarchical Python configurations to
describe an experiment end to end. A configuration specifies the model, data pipeline, training
strategy, evaluation protocol, and deployment interface, while inheritance
allows common settings to be reused across models, datasets, and embodiments.
After resolution, the configuration serves as the single source of truth for
component construction and runtime behavior. This design makes experiments
reproducible and allows training, evaluation, and inference to share consistent
data and model semantics without relying on separately maintained scripts.
The optional RL frontend composes these native model specifications with an
RLinf experiment configuration, preserving model semantics while allowing the
backend to own rollout and optimization settings.

\paragraph{A closed train--evaluate lifecycle.}
\fluxvla supports both standalone evaluation and an
\texttt{eval-after-train} workflow. When enabled, the checkpoint produced by
training is automatically passed to the evaluator defined by the same
experiment configuration. Evaluation begins as a separate stage after the
training environment has been released, preserving a clean execution boundary
while maintaining consistent model, data, and evaluation semantics. This
closed lifecycle reduces manual coordination, avoids configuration and
checkpoint mismatches, and makes reported results easier to reproduce.

\paragraph{A self-contained inference artifact.}
A common deployment failure in robot-learning systems is that a training
checkpoint contains only task-specific updates while implicitly depending on
the original pretrained weights, configuration files, or preprocessing assets.
Such hidden dependencies make checkpoints difficult to transfer and can cause
training and inference to use inconsistent model or data semantics.

The standard \fluxvla training path addresses this problem by packaging the
complete trained model state together with the resolved configuration and the metadata required for
inference. Evaluation and deployment can therefore reconstruct the trained
policy directly from the run artifact without locating the original
pretraining weights. This self-contained handoff reduces configuration and
version mismatches and provides a consistent interface between training,
simulation, serving, and physical-robot deployment, while leaving the software
environment and hardware-specific dependencies explicit.
The optional RL backend has a separate restoration lifecycle, described in
\cref{sec:rl-post-training}; its checkpoints are not automatically covered by
this native artifact guarantee.

\subsection{Data pipeline}
\label{sec:data-pipeline}

\begin{figure*}[t]
  \centering
  \includegraphics[width=\textwidth]{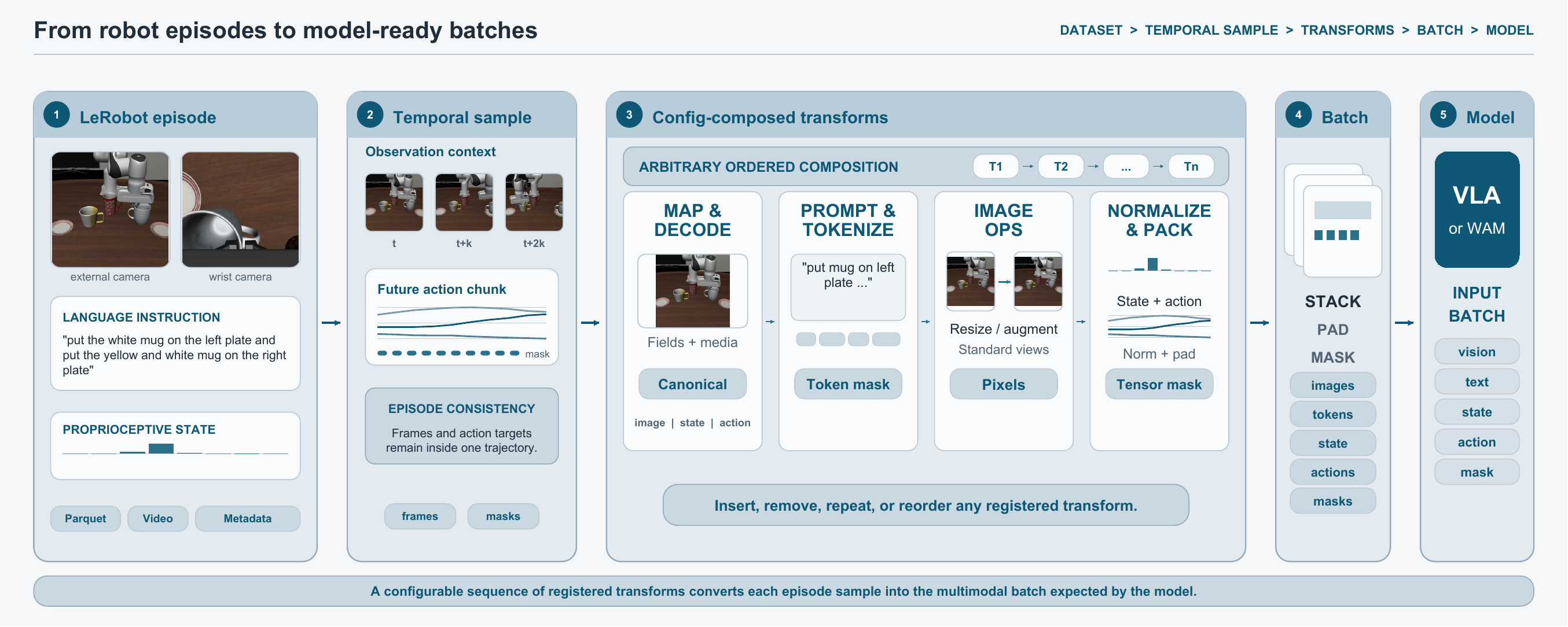}
  \caption{Configuration-driven data processing in \fluxvla.
  Offline trajectories and online observations share model-facing transforms
  and normalization semantics. Dataset ingestion, temporal sampling, and
  action recovery remain explicit stages.}
  \label{fig:data-pipeline}
\end{figure*}

The data pipeline has two responsibilities that are easy to conflate. First,
it must ingest trajectories produced by different simulators and robot
platforms. Second, it must present each model with the exact observation,
language, state, and action semantics assumed by that model. \fluxvla separates
these responsibilities into a storage-facing dataset layer and a
configuration-defined transform layer. Dataset classes recover temporally
consistent samples from an episode store; transforms then convert those
samples into a model-facing dictionary. This separation keeps physical storage
choices out of model code and makes differences in preprocessing visible in
the experiment configuration. As illustrated in \cref{fig:data-pipeline}, raw
sources are first converted to a unified Parquet representation compatible with
the LeRobot dataset format~\cite{lerobot2026}. A \texttt{ParquetDataset} then
owns an ordered list of
registered transforms that can be inserted, replaced, or rearranged entirely
through configuration. Online inference follows the same pattern: an
environment-specific dataset or adapter receives observations and executes its
own configured transform list to produce the same model-facing semantics.

\paragraph{Episode storage and ingestion.}
The principal offline representation is compatible with the LeRobot dataset
layout~\cite{lerobot2026}. Low-dimensional signals and episode indices are
stored in Parquet,
while camera observations are referenced from per-episode video files;
metadata records the feature schema, tasks, episodes, and per-feature
statistics. \fluxvla reads both the established LeRobot v2.1 metadata layout
and the v3 layout, whose tasks and episode metadata may themselves be stored as
Parquet files. This compatibility allows the data layer to use the Hugging Face
\texttt{datasets} implementation for indexed low-dimensional records without
requiring images to be expanded into the tabular store.

For data collected outside this representation, conversion is treated as an
explicit ingestion step. The provided ALOHA conversion path, for example,
turns HDF5 episodes containing joint state, optional end-effector pose, action,
and camera streams into LeRobot-compatible Parquet, metadata, and video assets.
The converter also makes otherwise implicit embodiment conventions---such as
camera names, arm ordering, gripper representation, and video frame rate---part
of the generated dataset. Dataset roots can optionally carry a FluxVLA content
version; when an expected version is configured, loading fails on a missing or
mismatched version rather than silently training on stale data.

\paragraph{Episode-aware temporal sampling.}
Robot trajectories contain temporal dependencies that cannot be captured by
treating individual rows as independent samples. \fluxvla therefore constructs
each training example from an episode-consistent temporal window containing the
current observation, a future action chunk, and, when required, multi-frame
visual context. Sampling never crosses episode boundaries, while masks prevent
padded observations or actions from contributing unintended supervision.
Temporal and camera dimensions are canonicalized before model-specific
preprocessing, allowing the same stored trajectories to support single-frame
VLAs, action-chunk policies, and video-based world--action models.

\paragraph{Canonical sample representation.}
Robot datasets often encode equivalent observations and actions using
dataset-specific schemas. \fluxvla resolves these differences through a
canonical sample representation that unifies visual observations,
proprioceptive states, language instructions, action targets, validity masks,
and optional metadata. This representation forms a stable interface between
data storage and model-specific preprocessing. Its extensible dictionary-based
design accommodates additional modalities, embodiment information, and
learning signals without changing unrelated components, thereby simplifying
the integration of heterogeneous datasets and policy families.

\paragraph{Statistics and normalization.}
\fluxvla derives normalization statistics from the associated dataset metadata
and applies them consistently to states and actions. Multiple normalization
strategies are supported to accommodate heterogeneous data distributions and
mixed continuous or discrete representations. The statistics used during
training are preserved with the resulting model artifact and reused during
evaluation and deployment. This coupling prevents discrepancies between
training-time preprocessing and inference-time action recovery, while allowing
different datasets and policy families to share the same data pipeline.

\paragraph{Multi-source sampling and batching.}
The outer dataset wrapper turns one or more finite episode datasets into the
iterable consumed by distributed training. It supports a single source,
concatenated sources with shared statistics, and grouped sources that retain
separate normalization domains. Samples are deterministically sharded over
both distributed ranks and data-loader workers; optional epoch-wise
reshuffling changes the order without duplicating a rank's shard. A balanced
variant defines a virtual epoch over multiple sources so that large datasets do
not automatically dominate small ones, with optional configured sampling
weights when the intended training mixture is not uniform.

The final collator makes batching behavior explicit. Fixed-shape numeric fields
are stacked, textual or diagnostic metadata is retained as lists, and
model-specific collators pad variable-length language or action-token sequences
while constructing attention masks and ignore labels. The resulting batch is
passed to the VLA without additional data-dependent logic in the training
loop. This keeps the runner agnostic to whether a batch originated from one
robot, several embodiments, or a mixture of model-specific preprocessing
paths.

\paragraph{Consistency between training and closed-loop inference.}
Simulation and real-robot execution begin from online observations rather than
Parquet rows, so they use environment-specific input adapters. Nevertheless,
the adapters converge to the same model-facing dictionary and reuse the same
registered prompt, image, state, and action transforms. Evaluation loads the
normalization statistics stored with the checkpoint, applies the configured
state normalization before prediction, and applies the inverse action
transform before sending commands to the environment or robot. Image-history
buffers and episode-reset signals are handled explicitly for temporal models.
Thus, training and deployment need not share the same storage backend, but they
share the semantic preprocessing contract that determines what the policy
actually observes and what its output means.

\subsection{Model architecture}
\label{sec:model-architecture}

\begin{figure*}[t]
  \centering
  \includegraphics[width=\textwidth]{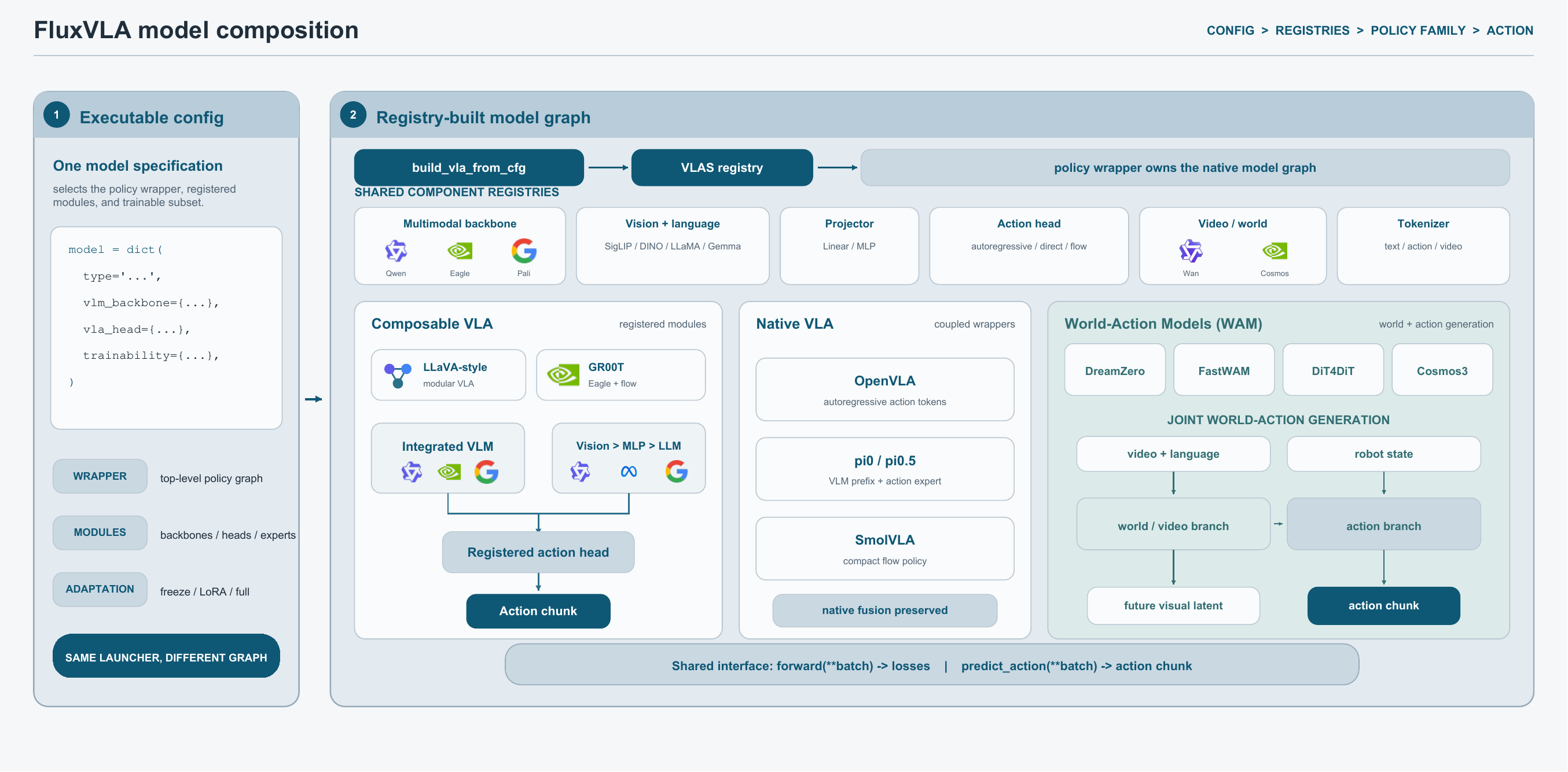}
  \caption{Configuration-driven model composition in \fluxvla.
  Registered backbones and action heads can be composed where compatible;
  native wrappers preserve tightly coupled policy architectures.
  The diagram shows model construction, not the execution lifecycle.}
  \label{fig:model-architecture}
\end{figure*}

The model layer is designed to preserve the native structure of substantially
different VLA algorithms while exposing a small interface to the rest of the
system. This distinction is important: forcing every policy into one internal
network topology would simplify the launcher at the cost of changing the
methods being reproduced. Instead, \fluxvla standardizes construction, batch
semantics, loss reporting, action prediction, and distributed wrapping, while
allowing each model family to retain its own multimodal fusion and action
generation procedure. \Cref{fig:model-architecture} presents this construction
boundary in the original configuration-to-model layout: an executable model
description selects registered components or a native policy wrapper, while
the execution-engine and deployment lifecycle are intentionally omitted.

\begin{table*}[t]
  \tablestyle
  \caption{Principal policy composition patterns represented in the \fluxvla
  model layer. The table describes shared engineering paths rather than
  claiming that the listed algorithms are architecturally identical.}
  \label{tab:model-patterns}
  \small
  \setlength{\tabcolsep}{4.5pt}
  \begin{tabularx}{\linewidth}{@{}P{0.22\linewidth}CCC@{}}
    \toprule
    Composition pattern & Representative integrations & Conditioning path & Action interface \\
    \midrule
    Autoregressive action tokens
      & OpenVLA
      & Vision encoder, projector, and causal language model
      & Token cross-entropy followed by action de-tokenization \\
    \addlinespace[3pt]
    VLM features with a replaceable head
      & LLaVA-style policies and GR00T
      & Integrated or decomposed VLM produces masked hidden features
      & Direct or flow-matching head produces continuous actions \\
    \addlinespace[3pt]
    VLM with an action expert
      & $\pi_0$, $\pi_{0.5}$, and SmolVLA
      & Visual--language prefix conditions state, time, and noisy-action tokens
      & Iterative flow integration produces an action chunk \\
    \addlinespace[3pt]
    Temporal or world--action model
      & DreamZero, DiT4DiT, FastWAM, and Cosmos3
      & Language and video latents are processed with temporal/action streams
      & Action objective, optionally coupled to dynamics or visual generation \\
    \bottomrule
  \end{tabularx}
\end{table*}

\paragraph{A configurable model graph.}
The common \texttt{BaseVLA} class defines slots for a vision backbone, a
language backbone, an integrated vision--language backbone, a modality
projector, and an action head. A model may use a decomposed path---for example,
a vision encoder whose patch features are projected into the embedding space
of a causal language model---or delegate multimodal fusion to an integrated
VLM. The optional action head consumes the fused representation together with
proprioceptive state and other conditioning variables. Each slot is built from
its own registry and configuration block, so replacing a visual encoder,
projector, VLM, or action head does not require a new training entry point.

This graph is intentionally permissive rather than artificially uniform.
Architectures such as $\pi_0$ contain a coupled language model and action
expert; world--action models may contain a video VAE, a text encoder, and
multiple diffusion-transformer streams; Cosmos3 performs modality packing
inside a mixture-of-transformers backbone. Such components remain owned by
their model-specific VLA class when their interaction is part of the original
algorithm. They still participate in the same registry, checkpoint, runner,
and inference contracts. Modularity therefore means that architectural
differences have explicit boundaries, not that all networks are reduced to an
interchangeable sequence of layers.

\paragraph{Unified training and prediction contracts.}
Different policy families often require model-specific inputs, objectives, and decoding procedures, forcing training and evaluation pipelines to accumulate specialized branches. \fluxvla addresses this fragmentation through a common model-facing contract: policies consume a canonical multimodal representation and produce standardized training objectives or executable action sequences. Architecture-specific validation, masking, fusion, and generation remain encapsulated within each policy, while environment-specific post-processing is handled by the corresponding adapter. Consequently, autoregressive policies, flow-based VLAs, and world--action models can share the same training and evaluation workflows without sacrificing their native algorithmic structure.

\paragraph{Multimodal representation and conditioning.}
For decomposed VLA models, the vision backbone converts one or more camera
views into patch features, and a linear or multilayer projector maps those
features to the language-model width. The resulting visual embeddings are
inserted into the language sequence with corresponding attention masks. An
integrated VLM instead returns a normalized backbone output containing hidden
states, a fused attention mask, and, when applicable, auxiliary losses. This
small output contract allows backbones such as Eagle, PaliGemma, Qwen-VL,
Florence, and SmolVLM to differ internally while remaining usable by a common
model-facing interface.

Proprioception and embodiment information are kept separate until the point at
which the target algorithm consumes them. A flow head may project the robot
state into a conditioning token, concatenate an embodiment embedding, and
cross-attend to VLM features. An action-expert architecture may instead append
state, noised-action, and time embeddings as a suffix to a cached
visual--language prefix. Temporal models receive canonical video tensors and
frame masks rather than independent images. In each case, the data pipeline
defines the physical meaning of the values, while the model defines how those
values enter its computation graph.

\paragraph{Autoregressive action generation.}
The OpenVLA path follows the discrete-action formulation of
OpenVLA~\cite{openvla2024}. A continuous action vector is quantized by an action
tokenizer and appended to the language target. If $z_k$ is the token assigned
to action component $k$, training minimizes masked next-token
cross-entropy,
\[
  \mathcal{L}_{\mathrm{tok}} =
  -\frac{1}{\sum_k m_k}
  \sum_k m_k \log p_\theta
  \!\left(z_k \mid \mathbf{o},\mathbf{l},z_{<k}\right),
\]
where $\mathbf{o}$ and $\mathbf{l}$ denote visual observation and language
instruction, and $m_k$ excludes non-action or padded targets. Inference uses
the causal model's generation path and converts the generated token IDs back
to continuous values. The implementation also reports token accuracy and the
$\ell_1$ error after de-tokenization, separating the optimized language-model
objective from a metric in the robot action space.

\paragraph{Continuous flow-matching policies.}
Continuous policies retain action chunks as real-valued tensors and learn a
velocity field between demonstrations and noise. Under the convention used by
the $\pi_0$ implementation, for a clean action chunk $\mathbf{a}$, noise
$\boldsymbol{\epsilon}$, and sampled time $t$, the interpolated input and target
velocity are
\[
  \mathbf{x}_t=(1-t)\mathbf{a}+t\boldsymbol{\epsilon},
  \qquad
  \mathbf{u}_t=\boldsymbol{\epsilon}-\mathbf{a}.
\]
The model predicts $\mathbf{v}_\theta(\mathbf{x}_t,t\mid
\mathbf{o},\mathbf{l},\mathbf{s})$, and training applies masked mean-squared
error over valid timesteps and action dimensions. At inference, sampling starts
from noise and numerically integrates the learned field toward a clean action
chunk. GR00T's feature-conditioned diffusion-transformer head,
$\pi_0$/$\pi_{0.5}$ action experts, and SmolVLA's compact expert use different
fusion and parameterization details~\cite{gr00t2025,pi02024,pi052025,smolvla2025};
\fluxvla preserves those details behind the shared loss and
\texttt{predict\_action} interfaces. The same reduction path can also apply
per-sample weights, allowing data-selection or reward-derived weighting to
alter behavioral-cloning emphasis without rewriting each optimizer loop.

\paragraph{Temporal and world--action models.}
The model boundary also accommodates policies whose visual input is a temporal
object rather than a set of current-frame features. DreamZero combines encoded
video latents, language, state, and actions under a joint flow-matching
objective~\cite{dreamzero2026}; its \fluxvla integration reports the visual
dynamics and action loss components separately. DiT4DiT couples a video
diffusion transformer with a separate action diffusion transformer conditioned
on visual dynamics and proprioceptive state~\cite{dit4dit2026}; its \fluxvla wrapper enforces the
corresponding video, state, action, and mask inputs through an explicit tensor
contract. FastWAM constructs video and action experts around a shared
mixture-of-transformers computation~\cite{fastwam2026},
while Cosmos3 packs text, visual latents, and action tokens into a multimodal
flow-matching sequence and can optionally optimize a visual objective alongside
the action objective. The shared runner only consumes their total loss and
action output; cache management, latent layout, repeated diffusion steps, and
auxiliary objectives remain local to the corresponding implementation.

\paragraph{Initialization and adaptation.}
Model construction is separated from pretrained-weight import. A VLA can load
\texttt{safetensors}, a directory of tensor shards, or a PyTorch checkpoint.
Because upstream projects frequently use different module names, configurable
name mappings translate external parameter paths to the \fluxvla module graph;
strict loading can be enabled when an exact mapping is required. Shape checks
and explicit mapping failures make partial or incompatible initialization
visible instead of silently redefining the model.

Trainability is likewise configuration controlled. Vision, language,
integrated VLM, and projector modules may be frozen independently; specialized
backbones can refine this policy for submodules that must remain trainable.
Full-parameter training and LoRA adaptation share the same dataset and runner
contracts. Finally, each VLA provides an FSDP wrapping policy aligned with its
transformer blocks and constituent modules. These mechanisms keep optimization
choices outside the semantic definition of the policy while ensuring that a
large heterogeneous model can still be initialized, fine-tuned, checkpointed,
and restored through the common system lifecycle.

\subsection{Simulation evaluation}
\label{sec:simulation-evaluation}

\begin{figure*}[t]
  \centering
  \includegraphics[width=\textwidth]{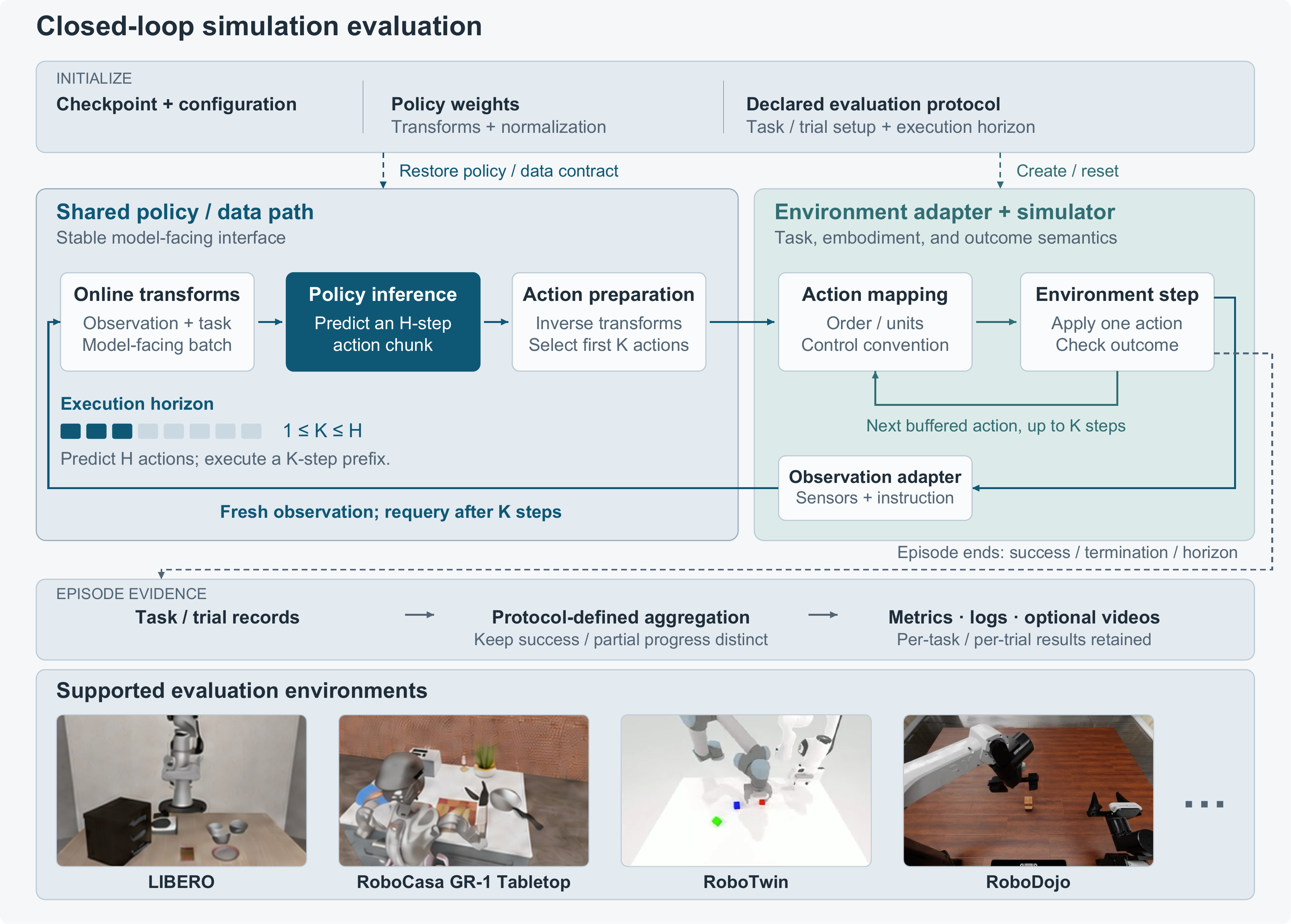}
  \caption{Closed-loop simulation evaluation in \fluxvla.
    Shared policy inference and data processing are separated from
    environment-specific interaction. The evaluator controls action-chunk
    execution and policy requerying, while retaining task- and trial-level
    evidence for aggregation. Scene images illustrate supported evaluation
    environments.}
  \label{fig:simulation-evaluation}
\end{figure*}

Simulation evaluation in \fluxvla reconstructs a trained experiment in closed
loop rather than invoking a model in isolation. Each rollout restores the
policy and preprocessing state, initializes an environment under an explicit
protocol, coordinates inference with action execution, and records episode
outcomes. Boundary mismatches can change measured performance without
changing model weights. The design therefore separates the shared policy and
data contracts from environment-specific interaction, which remains confined
to the evaluator and its adaptation logic.
Three properties define this boundary: faithful reconstruction of the policy
and evaluation setup from the selected checkpoint and associated artifacts;
isolation of simulator and embodiment semantics from model implementation;
and preservation of task-, trial-, and rollout-level evidence beyond
aggregate metrics. \Cref{fig:simulation-evaluation} illustrates this workflow,
and \cref{tab:simulation-evaluation} distinguishes shared responsibilities
from environment-dependent mechanisms. Concrete task sets, evaluation
budgets, and benchmark results are reported in the experimental evaluation
rather than defining the system interface here.

\begin{table*}[t]
  \tablestyle
  \caption{Shared evaluation contract and environment-dependent
  responsibilities. These are interface boundaries, not a claim that every
  integration implements identical scheduling, seeding, or metric features.}
  \label{tab:simulation-evaluation}
  \small
  \setlength{\tabcolsep}{5pt}
  \begin{tabularx}{\linewidth}{@{}P{0.18\linewidth}YY@{}}
    \toprule
    Concern & Shared evaluation contract & Environment-dependent responsibility \\
    \midrule
    Policy restoration
      & Resolve model, checkpoint, preprocessing assets, and normalization statistics
      & Select compatible embodiment or task-group settings; validate action and state conventions \\
    \addlinespace[3pt]
    Task and trial definition
      & Make the task set, trial budget, horizon, and reset protocol explicit
      & Instantiate environments and resolve initial states, layouts, randomization, and supported seed controls \\
    \addlinespace[3pt]
    Online input path
      & Transform observations and instructions into the model-facing batch
      & Extract sensor fields and task descriptions; manage episode-local observation history \\
    \addlinespace[3pt]
    Action execution
      & Schedule policy queries and apply inverse action transforms before stepping
      & Map dimensions, ordering, units, and control modes to the simulator interface \\
    \addlinespace[3pt]
    Outcomes and metrics
      & Preserve episode outcomes and aggregate under the declared protocol
      & Define success, termination, truncation, and any partial-progress metrics \\
    \addlinespace[3pt]
    Scheduling and evidence
      & Identify independent rollout units and retain structured results
      & Supply supported sharding, reproducibility controls, timing, and optional recording hooks \\
    \bottomrule
  \end{tabularx}
\end{table*}

\paragraph{Evaluation workflow reconstruction.}
An evaluator reconstructs the policy and evaluation setup associated with a
selected checkpoint from explicit configuration and associated artifacts,
avoiding hidden defaults. The configured execution path resolves the inference
model, online transform pipeline, inverse action transform, environment
protocol, and output configuration. An inference-oriented implementation may
replace its training counterpart for rollout
(\cref{sec:inference-acceleration}), provided checkpoint validation confirms
state compatibility. Dataset statistics restore the training-time
normalization contract. Simulator creation and stepping remain within the
evaluator, while model construction and invocation continue through the
common VLA interface. This separation avoids propagating simulator-specific
branches across model implementations.

\paragraph{Closed-loop rollout and execution horizon.}
At the beginning of an episode, the evaluator initializes the environment
according to the configured reset protocol and clears episode-local state
where required. It constructs an observation dictionary from sensor values
and the task instruction. The online data path applies the ordered transforms
in \cref{sec:data-pipeline}, and the resulting batch is passed to the policy.
For a policy predicting an $H$-step action chunk, the evaluator executes a
configured $K$-step prefix, $1 \leq K \leq H$, before requesting another
prediction; single-step execution is the special case $H=K=1$.
Inverse normalization and any embodiment-specific mapping precede simulator
execution. The evaluator checks environment-defined outcomes after each
transition and stops on the applicable success, termination, or horizon
condition. Smaller prefixes allow more frequent feedback, while longer
prefixes amortize inference cost. The schedule belongs to the evaluator,
not the policy architecture, and its configuration is part of the reported
evaluation protocol.

\paragraph{Environment and embodiment adaptation.}
Environment adaptation specifies how tasks are instantiated, initial states
or seeds are selected, observation fields are interpreted, and predictions
are converted to executable commands. These responsibilities may include
reset-time settling, temporal-history management, task-group statistics, or
partitioning actions across multiple controllers, depending on the
integration. The shared contract does not prescribe an action dimensionality,
robot morphology, or simulator API.
Action ordering, coordinate conventions, units, and normalization must remain
consistent with training. Checks on these semantics complement tensor-shape
validation: a shape-compatible command can still control the wrong joint or
use the wrong physical scale. Model-specific preprocessing remains in the
data path, while environment-specific command conversion remains at the
evaluator boundary. Extending evaluation to another environment therefore
requires compatible task, observation, action, and outcome handling rather
than changes to every supported policy.

\paragraph{Metrics, distributed execution, and auditable artifacts.}
An episode is identified by its task and trial configuration. Where supported,
independent episodes can be sharded across workers by episode for load balance
or by task to amortize initialization. Aggregation follows the declared
protocol rather than assuming that all tasks, groups, or settings have equal
weight. Full-task success and partial-progress measures remain separate;
additional metrics are reported only when defined by the environment and
recorded by the integration.
Reproducibility likewise depends on the available environment and inference
controls. Stored initial states, environment seeds, action-sampling seeds,
and scheduling configuration are recorded or fixed where supported, without
claiming identical determinism across simulators. Structured summaries,
per-task results, rank logs, and optional rollout videos connect aggregate
performance to the underlying episodes. Comparisons between checkpoints,
including supervised and RL-refined policies, require matched evaluation
conditions; a shared interface alone does not establish a controlled
comparison.

\subsection{Real-robot inference}
\label{sec:real-robot-inference}

Real-robot deployment introduces boundaries that are absent from offline
training and partially represented by simulation. Observations arrive from
independent sensor processes, commands must match an embodiment-specific
controller, and model latency competes directly with the robot control schedule.
\fluxvla addresses these difficulties by separating the policy-side inference
runner from the hardware-facing operator. The runner owns model restoration,
task state, model-facing observation history, and action-chunk prediction.
The operator acquires and synchronizes timestamped sensor streams and
transports command trajectories to the robot. Consequently, \fluxvla can reuse the same trained model and preprocessing
configuration while adapting only the robot-specific observation and actuation
interfaces.

\paragraph{A four-stage runner contract.}
\fluxvla defines real-robot inference as a four-stage control loop:
observation preprocessing, action prediction, action postprocessing, and
trajectory execution. The runner maintains deployment state, including control timing,
action chunk buffer, and optional temporal context, so the policy interface remains
focused on action generation. The same contract is used for both local and
remote inference: local execution runs the model on the robot client computer, whereas
remote execution sends observations to an accelerated server and receives
executable actions. This shared contract keeps observation and action semantics
consistent across simulation, local deployment, and remote serving.

\paragraph{Synchronized observation acquisition.}
Real-robot observations come from multiple sensor processes with different
rates and communication delays. Simply taking the latest message from each
stream can mix measurements from different physical states, producing an
inconsistent model input. \fluxvla therefore treats timestamp alignment as part
of the hardware-facing boundary rather than as a policy responsibility.

The operator synchronizes sensor messages by timestamp, maintains a bounded
buffer of valid observations, and monitors availability, freshness, and update
rate. Missing, stale, or poorly aligned inputs are reported explicitly before
they reach the policy. After synchronization, the observation is converted into
the canonical model-input format, while the runner maintains any fixed
observation history required by temporal policies. This keeps frame
synchronization and hardware-specific representations outside the policy while
preserving the preprocessing assumptions used during training.

\paragraph{Embodiment-aware action execution.}
Policy outputs follow the action representation used during training, whereas
robot controllers use embodiment-specific command spaces, dimension orderings,
control rates, and gripper conventions. The runner converts predictions into
robot-space actions and applies configured post-processing; the operator
handles hardware-specific command formatting, mode validation, and transport
for synchronous or asynchronous execution. This split follows
\cref{tab:runner-operator}, keeping model-side action recovery distinct from
controller-side command delivery. This design keeps robot-specific
conventions outside the policy, while low-level safety remains the
responsibility of the robot controller.

\paragraph{Remote GPU inference.}
The remote path decouples the latency-sensitive robot process from the model
runtime through a ZeroMQ request-reply service. The robot-side runner retains
user interaction and control-loop coordination, invoking the operator for
sensor acquisition, preparation poses, and command publication. It serializes raw observations with MessagePack or Protocol
Buffers: selected RGB arrays may be JPEG-compressed, other arrays are encoded
as NumPy payloads, and strings are sent directly. Each request contains an
observation and normalization key. The GPU server applies the online transforms, predicts and denormalizes
the actions, and returns serialized continuous actions.

The client profiles request serialization, ZMQ round-trip, server-side
inference, action deserialization, and total latency, and estimates
non-inference overhead from the difference between round-trip and inference
time. Because the ZMQ service uses plain TCP without built-in authentication
or encryption, it should be restricted to a trusted network or protected by an
SSH tunnel.

\paragraph{Safety boundary.}
The real-robot layer provides observation synchronization, dimensional and mode validation, preparation commands, rate-controlled publishing, trajectory cancellation, and graceful cleanup. Individual integrations may expose additional safeguards or control hooks; for example, the Tron2 operator forwards emergency-stop requests, while Franka can fall back from unavailable gripper action servers to a publisher interface. However, FluxVLA does not itself implement collision avoidance, workspace constraints, torque limits, or a certified emergency controller. These protections must be enforced by the robot SDK, low-level controller, and deployment procedure. A valid deployment therefore requires the dataset action convention, normalization statistics, runner action mode, operator command mode, active embodiment, and hardware safety envelope to be consistently aligned.

\subsection{Inference acceleration}
\label{sec:inference-acceleration}
\begin{figure*}[t]
  \centering
  \includegraphics[width=\textwidth]{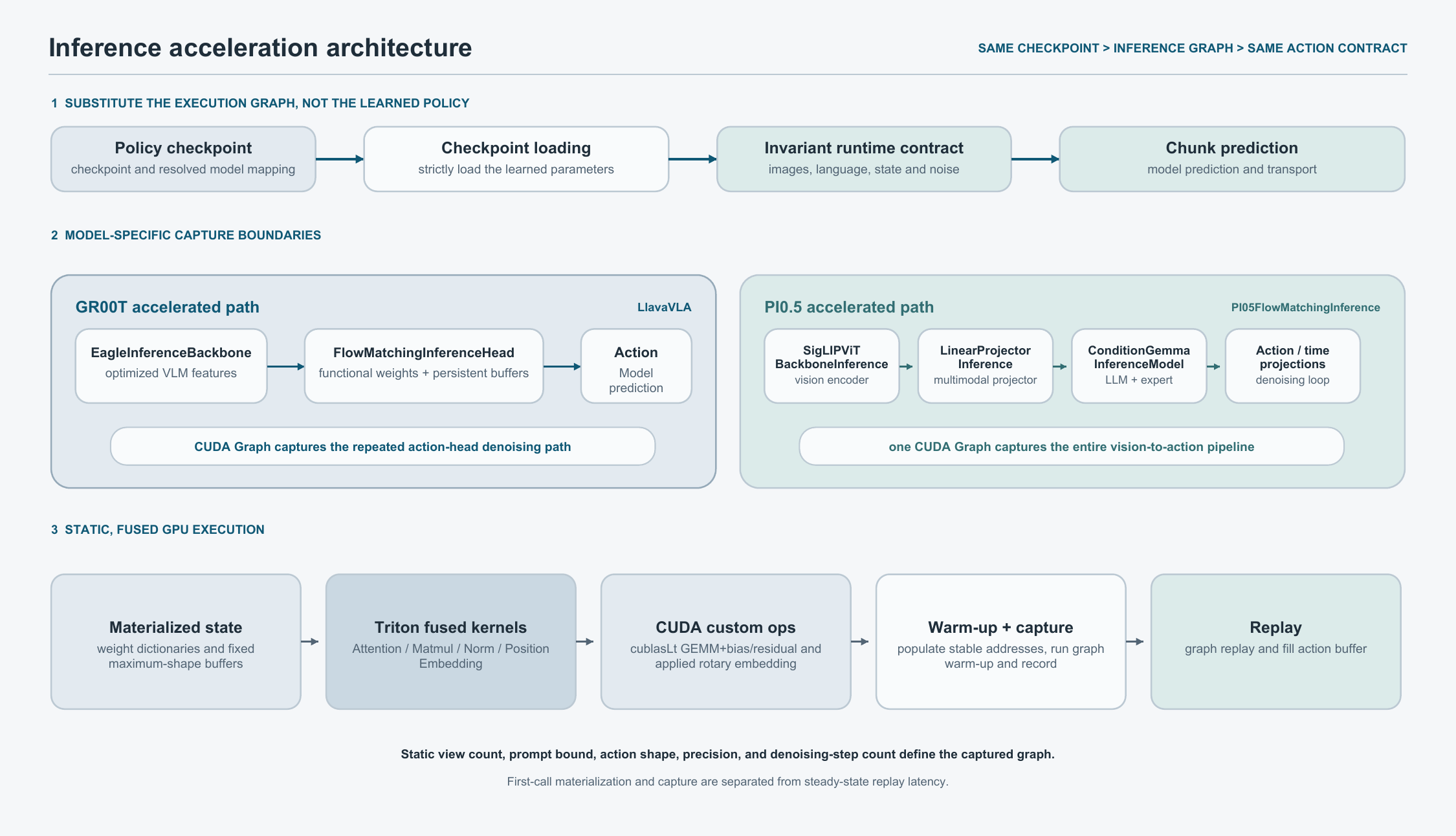}
  \caption{Inference acceleration in \fluxvla. GR00T captures the
  action-head path; $\pi_{0.5}$ captures the full vision--language--action
  pipeline. Both use persistent buffers and fused kernels for CUDA Graph
  replay while retaining the runner-facing action contract.}
  \label{fig:inference-acceleration}
\end{figure*}

VLA training implementations are designed to make experimentation reliable: they expose clear module boundaries, support gradient computation, and preserve architectural flexibility. At deployment time, however, the limiting factor shifts to action-prediction latency. Excessive inference delay can make manipulator execution appear discontinuous or sluggish, and can prevent the system from handling highly dynamic tasks. Inference acceleration is therefore a deployment requirement rather than a cosmetic optimization. \fluxvla addresses this by allowing the configuration to define an inference model that remains structurally compatible with the trained model while replacing selected modules, or the complete prediction path, with an inference-oriented implementation. The checkpoint remains unchanged: acceleration is a different execution graph over the same learned parameters, not a separately trained policy. \Cref{fig:inference-acceleration} summarizes the two code-level capture boundaries and the shared fused execution stack.

\begin{table*}[t]
  \tablestyle
  \caption{Layers of the accelerated inference path. The exact combination is
  model dependent; unsupported shapes or operators fall outside the static
  captured path rather than being silently approximated.}
  \label{tab:inference-acceleration}
  \small
  \setlength{\tabcolsep}{4.5pt}
  \begin{tabularx}{\linewidth}{@{}P{0.20\linewidth}CC@{}}
    \toprule
    Layer & Mechanism & Runtime effect \\
    \midrule
    Model substitution
      & Configuration selects inference-specific backbones, projectors, heads, or a complete VLA class
      & Retains the training checkpoint while removing training-oriented module paths \\
    \addlinespace[3pt]
    Functional execution
      & Weights and intermediate tensors are materialized into explicit dictionaries and fixed buffers
      & Reduces Python and \texttt{nn.Module} dispatch inside repeated denoising steps \\
    \addlinespace[3pt]
    Triton fusion
      & Fused normalization, projection, activation, attention, and position-embedding kernels
      & Avoids intermediate allocations and repeated global-memory traffic \\
    \addlinespace[3pt]
    CUDA operators
      & Specialized cublasLt GEMM--bias/residual and rotary-embedding kernels
      & Uses fused low-level implementations for selected hot operations \\
    \addlinespace[3pt]
    CUDA Graph replay
      & Warm up and capture a fixed-shape GPU execution sequence once
      & Replaces repeated host-side kernel submission with graph replay \\
    \bottomrule
  \end{tabularx}
\end{table*}

\paragraph{Inference-model substitution.}
Model selection is performed before checkpoint restoration. For GR00T-style
models, the optimized path retains the top-level \texttt{LlavaVLA} composition
but substitutes an \texttt{Eagle\allowbreak Inference\allowbreak Backbone} and
\texttt{FlowMatching\allowbreak Inference\allowbreak Head} for their training counterparts. For
$\pi_{0.5}$, the \texttt{PI05\allowbreak FlowMatching\allowbreak Inference} class replaces the complete
vision--language and action-generation pipeline. Configured name mappings align
the checkpoint parameter paths with these inference modules, and strict state
loading exposes a missing or incompatible mapping. This arrangement keeps
optimization local to the model implementation: the simulation and robot
runners continue to call the same \texttt{predict\_action} interface.

\paragraph{Kernel fusion and functionalization.}
The optimized paths target small operations that become expensive when they
are repeated across many transformer layers and denoising iterations. The
Triton normalization and matrix-multiplication operators, together with parts
of the optimized inference backbone, build on the open-source Realtime-VLA
implementation~\cite{realtimevla2025}. The operator library includes fused
residual addition and normalization,
adaptive LayerNorm and RMSNorm, QKV projection with rotary embedding, gated
MLP projections with SiLU, matrix multiplication with bias and activation, and
in-place position-embedding addition. These kernels combine operations that
would otherwise allocate intermediate tensors and launch separate GPU kernels.

Selected hotspots are implemented as custom CUDA extensions when direct use of
vendor primitives or a specialized memory layout is advantageous. The
cublasLt-based matrix multiplication path fuses GEMM, bias, and an optional
residual into one operation and accepts a preallocated output tensor. Separate
rotary kernels construct Gemma sine/cosine embeddings and apply rotary position
encoding to query and key tensors. Above these primitives, functional atomic
operations explicitly receive weights, buffers, and output locations rather
than relying on dynamically allocated intermediate module outputs.

\paragraph{Static buffers and CUDA Graphs.}
CUDA Graph replay requires stable memory addresses and a fixed execution
topology. The accelerated models therefore materialize weights and allocate
buffers for the maximum configured image, prompt, encoder, decoder, action,
and denoising dimensions. Runtime inputs are copied into these persistent
buffers; valid prompt length and related masks indicate which portion contains
data without changing the allocated shape. The first prediction performs
operator warm-up, synchronizes the device, captures the graph, and stores it.
Subsequent predictions update the input and noise buffers and replay the same
graph.

The scope of capture differs between the two principal implementations.
\texttt{FlowMatching\allowbreak Inference\allowbreak Head} functionalizes and captures the repeated
GR00T action-head computation after receiving VLM features. In contrast,
\texttt{PI05\allowbreak FlowMatching\allowbreak Inference} manually unrolls the vision encoder,
language-conditioned transformer encoder, action expert, and denoising loop
into one graph. Capturing the full $\pi_{0.5}$ pipeline avoids a host boundary
between the visual--language prefix and repeated action decoding, but also
requires fixed camera count, maximum prompt length, chunk length, precision,
and number of denoising steps.

\paragraph{Numerical and configuration contract.}
An optimized backend is expected to preserve the policy's external behavior:
it accepts the same transformed observation semantics and returns the same
continuous action layout. It need not reproduce every floating-point operation
bit for bit. Fused kernels, bfloat16 arithmetic, and reordered reductions can
produce small differences from eager PyTorch, so backend validation should
compare action-space error and trajectory or task behavior in addition to
model-loading success. The training model remains the reference path, and the
inference model is an explicit configuration choice rather than an automatic
global replacement.

Static capture also creates constraints that must remain visible. Inputs cannot
exceed the allocated prompt or sequence bounds, and changes to the number of
views, action horizon, action width, denoising-step count, or model structure
may require rematerialization and recapture. The first-call warm-up and graph
recording cost must be excluded or reported separately from steady-state
latency. Likewise, model frequency is not end-to-end control frequency: image
acquisition, preprocessing, host--device copies, remote transport,
denormalization, and robot command scheduling remain outside some captured
regions. For this reason, \fluxvla exposes acceleration through the same runner
and profiling boundaries used by deployment rather than treating an isolated
kernel benchmark as the complete runtime result.

\subsection{Real-Time Chunking}
\label{sec:rtc}

Action chunking reduces how often a VLA must be queried, but it introduces a
temporal coordination problem. Suppose a policy predicts
$\hat{\mathbf{A}}^{i}=(\hat{\mathbf{a}}^{i}_0,\ldots,
\hat{\mathbf{a}}^{i}_{H-1})$ and the robot begins executing it while the next
observation is processed. When prediction and execution overlap, part of a
newly generated action chunk may already be outdated by the time it becomes
available. The execution layer therefore schedules only temporally valid
actions, while RTC further conditions the next prediction on the portion of
the previous chunk that remains active.

Timing-aware scheduling prevents stale commands from being replayed, but it
does not ensure continuity between successive predictions. Independently
generating the new chunk can still produce an abrupt transition even when both
chunks are individually plausible. Real-Time Chunking (RTC) treats temporally
overlapping actions from the previous chunk as an inpainting target: commands
guaranteed to execute are fixed, while later overlapping actions receive
decaying consistency guidance~\cite{rtc2025}.

In this report, \textbf{Training-time RTC} and \textbf{Test-time RTC} name two
complete routes rather than two mandatory stages of one procedure. Training-time
RTC pairs simulated-delay conditioning during training with prefix conditioning
during inference~\cite{trainingtimertc2025}. The original Test-time RTC method
leaves model training unchanged and applies vector--Jacobian-product (VJP)
guidance during inference~\cite{rtc2025}. \fluxvla additionally implements a
lower-cost direct approximation to that guidance.

\begin{table*}[t]
  \tablestyle
  \caption{Training- and inference-stage mechanisms of the RTC routes
  implemented in \fluxvla.}
  \label{tab:rtc-modes}
  \small
  \setlength{\tabcolsep}{7pt}
  \begin{tabularx}{\linewidth}{@{}P{0.22\linewidth}CC@{}}
    \toprule
    RTC route & Training stage & Inference stage \\
    \midrule
    Training-time RTC
      & \textbf{Simulated-delay conditioning.} Expose a clean known-action
        prefix during training~\cite{trainingtimertc2025}
      & \textbf{Prefix conditioning.} Keep the aligned prefix fixed throughout
        denoising \\
    \addlinespace[3pt]
    Test-time RTC (VJP)
      & No RTC-specific training
      & \textbf{VJP guidance.} Differentiate the prefix-consistency error
        through the denoiser~\cite{rtc2025} \\
    \addlinespace[3pt]
    Test-time RTC (direct)
      & No RTC-specific training
      & \textbf{Direct guidance.} Add the weighted prefix error directly to
        the predicted velocity \\
    \bottomrule
  \end{tabularx}
\end{table*}

\paragraph{Runtime alignment of successive chunks.}
RTC aligns the previous raw action chunk with the time of the next query. Let
$t_{\mathrm{ref}}$ denote the time origin of the previous chunk's first sample
and $\Delta t$ the command period. The elapsed offset is
\[
  \delta = \frac{t_{\mathrm{query}}-t_{\mathrm{ref}}}{\Delta t},
\]
which may be fractional. The runner therefore linearly resamples the remaining
trajectory from $\delta$ and clamps the requested prefix length $L$ to the
available suffix. The resulting prefix refers to the commands that remain
relevant on the control timeline rather than to stale array positions. The
execution horizon controls when a new query is issued, whereas $L$ controls how
much overlap conditions that query.

\paragraph{Training-time RTC.}
Its training-stage mechanism is \textbf{simulated-delay
conditioning}~\cite{trainingtimertc2025}. For each
batch element, the implementation samples a simulated delay $d$, interpreted as
the length of a known action prefix. The first $d$ action positions are therefore
the known-prefix positions. Let
$m_j$ denote the validity mask at action position $j$. The known-prefix
positions satisfy $j<d$; their per-position diffusion or flow time is replaced
by the model's clean-time value, and their reconstruction loss is masked:
\[
  t_j' =
  \begin{cases}
    t_{\mathrm{clean}}, & j<d,\\
    t, & j\geq d,
  \end{cases}
  \qquad
  \tilde m_j = m_j\,\mathbf{1}_{\{j\geq d\}}.
\]
This construction exposes the model to a clean known prefix while asking it to
predict only the unknown suffix. The implementation accounts for differing
time conventions: the clean endpoint is $1$ for the GR00T flow head and $0$
for the $\pi_0$/$\pi_{0.5}$ convention. With a configured maximum delay $D$,
the implementation samples $d\in\{0,\ldots,D-1\}$; the upper bound is
exclusive. The simulated-delay distribution and exponential temperature are
likewise configuration parameters rather than fixed properties of the training
runner.

Its inference-stage mechanism is \textbf{prefix conditioning}. It initializes the
first $L$ positions with the aligned previous actions, restores them after every
denoising update, and provides a clean-time encoding for those positions.
Reapplying the constraint after the final step ensures that the returned overlap
exactly matches the prefix. Prefix conditioning remains executable without the
matched training procedure, but here \textbf{Training-time RTC} denotes the paired
use of simulated-delay conditioning and prefix-conditioned inference.

\paragraph{Test-time RTC.}
Test-time RTC leaves model training unchanged and applies RTC guidance only
during inference. It defines a position weight $w_j$ that is one over the
locked prefix and decays toward zero over a configurable region after the
boundary. Given a current noisy action $\mathbf{x}_t$, predicted velocity
$\mathbf{v}_t$, and implied clean estimate
$\hat{\mathbf{x}}_{\mathrm{clean}}$, the method forms a prefix error against
the aligned previous actions. The clean endpoint follows the model's time
convention rather than always occurring at $t=1$.

The original method differentiates this error through the denoiser and applies
the resulting vector--Jacobian product as a velocity correction~\cite{rtc2025}.
This incorporates the denoiser's local Jacobian but enables gradients during
inference and adds compute and memory cost. \fluxvla also provides a lower-cost
direct approximation that adds the weighted prefix error to the predicted
velocity without differentiating through the denoiser. This direct path is an
implementation-specific simplification, not a method attributed to the cited
work. Exponential, linear, constant-one, and zero decay schedules are available
for the prefix weights, together with a maximum guidance weight.
Training-time RTC and Test-time RTC remain alternative routes: a single query
uses either prefix conditioning or guidance, not both.

\paragraph{Interaction with accelerated inference.}
RTC must not invalidate the fixed-shape assumptions of CUDA Graph replay. The
accelerated GR00T head therefore allocates prefix-action, prefix-mask, and
prefix-timestep buffers before capture. Each query clears and fills the relevant
prefix region in place, while the captured graph always executes over the same
full shapes. \texttt{PI05FlowMatchingRTCInference} similarly extends the
whole-pipeline captured graph with prefix-aware decoder attention and in-place
prefill buffers. Both accelerated implementations currently support the
prefix-conditioning mechanism used by Training-time RTC. Both the original VJP
guidance and the direct approximation remain on the eager model paths: their
dynamic corrections are not part of the captured execution graph, and the VJP
method additionally requires gradients during inference.

Both RTC routes couple runner-side temporal alignment with a model-side
denoising intervention. Training-time RTC adds simulated-delay conditioning
during training and enforces the aligned prefix during inference. Test-time RTC
instead applies guidance to an ordinarily trained flow policy. In both cases, a
prefix without temporal alignment may be stale, and faster inference reduces
but does not by itself remove cross-chunk discontinuity.

\subsection{Trajectory Post-processing}
\label{sec:trajectory-postprocessing}

RTC improves temporal consistency while an action chunk is being generated,
but a prediction that is consistent with its predecessor is not necessarily
ready for direct execution. A raw robot-space trajectory may still contain
local oscillations, imply excessive velocity, acceleration, or jerk, or meet
the previous chunk with an undesirable change in higher-order motion. These
properties depend on the robot controller and command rate rather than only on
the policy distribution. \fluxvla therefore exposes trajectory
post-processing as an optional runner-side stage between action
denormalization and command execution.

The post-processing interface receives a timestamped trajectory containing
positions and, when available, velocities and accelerations. It operates only
on configured continuous degrees of freedom; discrete or mode-like channels
such as gripper and suction commands remain outside the continuous optimizer
and retain their embodiment-specific handling. The result preserves the
original action layout and horizon so that the downstream operator contract is
unchanged. This placement also keeps execution constraints independent of the
VLA architecture and applies equally to locally and remotely predicted action
chunks. The configuration separates three decisions: how the trajectory is
generated, what terminal behavior is desired, and whether an active previous
trajectory must be stitched to the new one. \Cref{tab:trajectory-postprocessing}
summarizes these orthogonal axes rather than treating their individual choices
as interchangeable methods.

\begin{table*}[t]
  \tablestyle
  \caption{Orthogonal configuration axes of trajectory post-processing. A
  deployment selects one backend and one execution mode, then independently
  enables stitching when cross-chunk boundary alignment is required.}
  \label{tab:trajectory-postprocessing}
  \small
  \setlength{\tabcolsep}{4.5pt}
  \begin{tabularx}{\linewidth}{@{}P{0.22\linewidth}CCC@{}}
    \toprule
    Configuration axis & Choices & Responsibility & Selection criterion \\
    \midrule
    Trajectory-processing backend
      & Joint-separable MPC or Ruckig filter
      & Convert a policy reference into a trajectory satisfying the selected higher-order motion limits
      & Horizon-level tracking fidelity versus lower post-processing latency and reference lag \\
    \addlinespace[3pt]
    Execution mode
      & Tracking or settle
      & Determine whether the processed trajectory should preserve rolling motion or approach its final target and rest
      & Asynchronous repeated replanning versus synchronous execution of a self-contained segment \\
    \addlinespace[3pt]
    Boundary handling
      & Stitching enabled or disabled
      & Reuse a time-aligned prefix from the previously processed trajectory before solving the new suffix
      & Whether a new chunk replaces a trajectory that is still being executed \\
    \bottomrule
  \end{tabularx}
\end{table*}

\paragraph{Joint-space finite-horizon optimization.}
The joint-space model predictive control (MPC) backend treats the denormalized
policy output $\hat{\mathbf q}_{0:H-1}$ as a reference trajectory. For each
continuous joint, it optimizes position, velocity, acceleration, and jerk over
the fixed action horizon. In a simplified vector notation, the objective is
\[
  \begin{aligned}
    \min_{\mathbf q,\mathbf v,\mathbf a,\mathbf j}\quad
    & w_{\mathrm{trk}}\sum_{t=0}^{H-1}
      \lVert\mathbf q_t-\hat{\mathbf q}_t\rVert_2^2 \\
    & {}+ \lambda\lVert\mathbf z\rVert_2^2 \\
    & {}+ w_{\mathrm{term}}\lVert\mathbf q_{H-1}-
      \hat{\mathbf q}_{H-1}\rVert_2^2 \\
    & {}+ w_{\mathrm{stop}}\lVert\mathbf v_{H-1}\rVert_2^2,
  \end{aligned}
\]
where $\mathbf z$ collects the optimized trajectory variables. The tracking
term is active in both modes, while the terminal-position and terminal-velocity
terms provide the soft settling bias when requested. The optimization obeys a
discrete triple-integrator model,
\[
  \begin{aligned}
    \mathbf q_{t+1} &= \mathbf q_t+\Delta t\,\mathbf v_t, \\
    \mathbf v_{t+1} &= \mathbf v_t+\Delta t\,\mathbf a_t, \\
    \mathbf a_{t+1} &= \mathbf a_t+\Delta t\,\mathbf j_t,
  \end{aligned}
\]
with the initial state anchored to the aligned execution state and box
constraints on velocity, acceleration, and jerk. The resulting quadratic
program is separable across joints and is solved with OSQP~\cite{osqp2020}.
The \fluxvla backend reuses one solver per joint and executes the independent
problems in parallel. This is an execution-side
kinematic trajectory optimizer: it does not model coupled robot dynamics,
torque, collision, or workspace constraints, which remain outside its
guarantee.

\paragraph{Jerk-limited filtering.}
Ruckig provides online trajectory generation under velocity, acceleration, and
jerk constraints~\cite{ruckig2021}. The \fluxvla backend uses it as a second
realization of the post-processing contract, advancing a feasible trajectory
toward successive policy targets rather than optimizing tracking error over the
complete horizon.
This construction makes higher-order bounds direct and computationally light,
but it can introduce reference lag because each update prioritizes feasible
motion from the current kinematic state. In tracking mode the fixed-length
result follows the rolling command stream. In settle mode the generator may
continue beyond the final reference to approach rest and then resample the
result to the policy horizon. Because temporal resampling changes derivatives,
limits must be verified again whenever an extended settle trajectory is
compressed to a fixed output length.

\paragraph{Current backend scope and extensibility.}
Both backends operate independently on each selected joint. They share the
chunk timestamps and horizon, but do not introduce cross-joint costs or coupled
constraints. This joint-separable design keeps the optimization small, permits
parallel execution, and is sufficient in the evaluated setting to obtain a
practical balance between reference precision and smoothness when velocity,
acceleration, and jerk bounds are chosen for the robot. Joint MPC favors closer
tracking of the policy reference, whereas Ruckig favors lower processing
latency and direct jerk-limited generation.

Joint separability is a property of the provided backends, not a restriction of
the post-processing boundary. A new implementation may instead optimize
coupled joints, operate in Cartesian space, enforce embodiment-specific
constraints, fit splines, or apply another filtering rule. To reuse the runner
and operator paths, it must consume the timestamped robot-space trajectory and
selected action channels, preserve the declared action layout, and return a
trajectory whose timing and horizon are explicit. Backend-specific parameters
and validity checks remain visible in the post-processing configuration.

\paragraph{Cross-chunk stitching.}
Post-processing retains the previously processed trajectory together with its
start time and command period. When a new chunk arrives during asynchronous
execution, the runner samples that trajectory at the timestamps assigned to
the beginning of the new chunk. The first $S$ samples are copied as a stitch
prefix, and optimization begins from the position, velocity, and acceleration
at the stitch boundary. The assembled result consists of this time-aligned
bridge followed by the newly optimized suffix. Thus stitching follows the
trajectory scheduled for execution rather than treating each policy output as
an isolated motion segment.

\paragraph{Relationship to RTC.}
RTC and trajectory post-processing are sequential rather than competing
mechanisms. RTC acts inside the policy query: it conditions the next normalized
action chunk on the unexecuted part of the previous raw prediction. The result
is then denormalized and passed to trajectory post-processing, which applies
joint-space bounds and, when enabled, stitches it to the previously processed
trajectory. The operator executes only this final robot-space trajectory.

This ordering gives the two stages distinct guarantees. RTC reduces the change
introduced when the policy generates a new chunk, but does not bound velocity,
acceleration, or jerk after denormalization. Post-processing regularizes the
executable command, but does not change how the policy represents or generates
the unknown suffix. Either stage can be used alone. When both are enabled, RTC
provides a more consistent reference for the postprocessor, while stitching
connects that reference to the trajectory actually sent to the robot. The
runner includes prediction and post-processing delay when discarding stale
samples before execution.

\paragraph{Configuration and operational boundary.}
The runner selects the backend, tracking or settle mode, continuous action
indices, stitch length, motion limits, and objective weights through a
post-processing configuration block. Tracking with stitching is the natural
choice for rolling asynchronous inference, whereas settle mode is intended for
synchronous execution of a self-contained segment. These settings are part of
the embodiment contract rather than universal model hyperparameters: the same
numerical limit can have different physical meaning under joint-position,
end-effector, or whole-body control. The downstream robot controller must
retain clipping, collision checking, emergency stopping, and other safety
interlocks; trajectory post-processing improves command regularity but is not
a certified safety controller.

\input{sections/03_rl_post_training}

%% file: sections/03_rl_post_training.tex
\subsection{Reinforcement learning post-training}
\label{sec:rl-post-training}

Supervised fine-tuning provides a useful policy initialization, but a fixed
demonstration set cannot cover every state encountered during execution.
Online RL offers a complementary route: collect experience with the current
policy and optimize it using environment feedback. \fluxvla supports this
transition through an optional integration with RLinf~\cite{yu2026rlinf,rlinfvla2025},
allowing existing flow-based policies to enter an interactive learning loop
without duplicating their model and preprocessing implementations.

\paragraph{Separation of policy and optimization.}
\fluxvla retains responsibility for model construction, initial checkpoint
loading, observation processing, and action conversion. RLinf manages
environment workers, rollout collection, distributed optimization, weight
synchronization, and training checkpoints. Rollout and learner workers
construct the same policy through a shared registry, while benchmark adapters
remain separate from model code. This division reuses the native policy
implementation without requiring the RL backend for ordinary supervised
training or inference.

\paragraph{From flow prediction to policy updates.}
A flow policy produces actions through iterative denoising, but this process
does not directly expose a tractable marginal action likelihood for policy
optimization. Following the Flow-SDE approach used in
$\pi_{\mathrm{RL}}$~\cite{pirl2025}, training rollouts introduce a Gaussian
transition at one randomly selected denoising step and retain the native ODE
updates elsewhere. The rollout records the sampled latent trajectory and its
transition log probabilities. During learning, the policy recomputes those
probabilities for the same stored transition rather than sampling a new
trajectory. These quantities describe denoising transitions, not the exact
marginal likelihood of the final robot-space action; the construction is
therefore specific to compatible flow policies.

\paragraph{Closed-loop optimization.}
A value head summarizes the policy's observation representation to estimate
returns. Environment rewards and value estimates support generalized
advantage estimation~\cite{gae2015} and clipped PPO
updates~\cite{ppo2017}. The current recipes freeze the multimodal prefix
encoder and optimize the action expert, associated action-generation
projections, and value head. RLinf alternates rollout collection with
FSDP2-sharded learner updates in synchronous rounds and synchronizes the updated policy before the
next collection phase. Stored latent samples and probability computations
retain full precision to reduce discrepancies between rollout and learner
computations; sharding boundaries respect each model's actual execution path.

\paragraph{Benchmark and model extensibility.}
The integration includes recipes for $\pi_{0.5}$ on LIBERO-10 and the
RoboTwin \texttt{adjust\_bottle} task. The paired RL evaluations in this report
focus on SmolVLA on LIBERO-10 and ten RoboTwin Clean tasks
(\cref{tab:rl-post-training-results}); no quantitative RL results for
$\pi_{0.5}$ are reported. Configuration coverage should therefore not be
read as evidence of measured RL gains for every integration. Adapters reuse each
policy's image, language, state, and normalization conventions, applying
embodiment-specific action conversion only after prediction. Additional
compatible flow policies can reuse the sampling and optimization interface
while supplying their own observation encoding and velocity computation.
Autoregressive or otherwise incompatible policies require an appropriate
likelihood interface rather than reusing the flow formulation unchanged.

\paragraph{Evaluation and artifact boundary.}
Evaluation uses the native ODE prediction path without the additional
training-time stochastic transition, enabling a matched comparison between
the initial supervised policy and its RL-refined counterpart. The RL
restoration path currently reconstructs the policy from its initial
supervised checkpoint and accompanying preprocessing assets before loading
the RL state. It is therefore distinct from the self-contained native
training artifact in \cref{sec:overall-design}. The integration establishes
the policy--backend interface; task-level gains require the paired evaluation
defined in \cref{sec:rl-results}.

%% file: sections/04_core_abstractions.tex
\section{Extensibility through Stable Contracts}
\label{sec:abstractions}

\Cref{sec:overview} described how a complete \fluxvla experiment moves
from data to training, evaluation, and deployment. This chapter instead
examines the interfaces that allow that workflow to be extended without
rewriting it. The central design choice is to stabilize the \textbf{boundaries}
between components rather than impose one implementation on every policy or
robot. A dataset may change its storage backend, a VLA may replace its action
decoder, and a robot may use a different middleware stack, provided that each
component continues to satisfy the contract at the boundary it crosses.

A contract in this context has four parts: a construction rule, an exchanged
data structure, lifecycle ownership, and an explicit extension point.
Registries and configuration determine how an object is constructed. Sample
and output dictionaries define what adjacent components exchange. Runners
determine when resources are initialized, used, checkpointed, and released.
Operators isolate the communication and actuation details that should not
enter the policy implementation. These contracts are intentionally narrower
than semantic compatibility: the fact that two objects can be constructed
from the same registry does not guarantee that their tensor shapes, temporal
semantics, or optimization assumptions agree. The configuration author is
responsible for selecting compatible components, while the framework makes the
selection visible and localizes the required adaptation.

\subsection{Abstraction hierarchy and extension boundaries}
\label{sec:abstraction-hierarchy}

The abstraction hierarchy follows the ownership structure of an experiment.
A launcher constructs a dataset and a runner. The dataset owns its ordered
transform pipeline; the runner owns the model, collator, metric, optimizer,
scheduler, and optional evaluator. The model in turn owns its vision,
language, integrated VLM, projector, and action-head modules. During physical
execution, an inference runner owns policy-side temporal logic and delegates
hardware I/O to an operator. This ownership hierarchy is more than an
implementation detail: it identifies the component that must validate a
nested configuration and the component whose lifecycle controls the nested
object.

This structure creates two kinds of extension boundary. A \textbf{horizontal}
boundary replaces one implementation within a family, such as one image
transform with another or one evaluator with another. A \textbf{vertical}
boundary introduces a new owner that composes existing lower-level modules,
such as a new VLA class that combines a registered VLM backbone with an action
head. Horizontal replacement is usually configuration-only when the semantic
contract is unchanged. Vertical extension normally requires a small amount of
Python code because the owner must define how its children interact.

The hierarchy also prevents configuration from becoming an unstructured
dependency injection mechanism. Nested modules are not built globally and then
passed to arbitrary consumers. Instead, each owner invokes the category-specific
builder for the children it understands. For example, a dataset interprets its
transform list in order, a VLA decides how backbone features enter an action
head, and a runner decides when to create the optimizer. This keeps semantic
knowledge near the code that can validate it and limits the effect of a new
integration on unrelated workflows.

\subsection{Configuration and construction contract}
\label{sec:configuration-contract}

Robotic learning systems often couple component selection with task-specific
control logic, making it difficult to reuse implementations or introduce new
models without modifying the overall workflow. \fluxvla addresses this problem
through registry-based construction: data processing, model, optimization,
evaluation, and deployment components are exposed through a common interface
and selected declaratively from configuration.

A unified hierarchical configuration describes both the component graph and
the relationships among its parts. The framework constructs the requested
modules recursively, allowing model backbones, action heads, preprocessing
operations, training strategies, and execution backends to be replaced or
recombined without introducing model-specific branches into the launcher.
Invalid or unavailable components are detected during construction, making
integration errors explicit rather than silently falling back to unintended
behavior.

This modularity does not imply that arbitrary components are automatically
compatible. Components must still agree on their semantic contracts, including
input fields, feature dimensions, temporal horizons, and action
representations. The configuration therefore serves as an executable
specification of the complete system rather than merely a collection of
hyperparameters.

By preserving the resolved configuration together with the trained model,
\fluxvla can reconstruct the same component composition and preprocessing
semantics during evaluation and deployment. This design improves
reproducibility while reducing inconsistencies between training, simulation,
serving, and physical-robot execution.

\subsection{Data contract}
\label{sec:data-contract}

Robot datasets and online observations often differ in storage layout, field
names, sensor organization, and action representation. Directly exposing these
differences to the model would couple each policy to a particular dataset or
deployment platform. \fluxvla instead defines a semantic, dictionary-based
boundary between the data and model layers.

Both offline trajectories and live observations are converted to this boundary
through configurable transform sequences. The data layer constructs the
required temporal context while respecting episode boundaries, and the
transforms progressively canonicalize, decode, normalize, augment, and encode
the sample. Each policy can therefore request the modalities and
representations it requires without imposing a single tensor layout on all
model families.

The contract is intentionally extensible. New observations, embodiment
descriptors, validity information, or auxiliary supervision can be introduced
without changing unrelated components, while explicit batching rules prevent
fields from being silently discarded or interpreted incorrectly. Models remain
independent of the original storage format and whether samples originate from
a single source or a mixture of datasets.

Training and inference may use different acquisition paths, but they preserve
the same model-facing semantics and preprocessing conventions. In particular,
normalization, modality ordering, and action interpretation remain coupled to
the trained policy. This shared contract reduces discrepancies between offline
training, simulation evaluation, remote serving, and physical-robot
deployment.

\subsection{Model contract}
\label{sec:model-contract}

VLA architectures differ substantially in how they combine visual, linguistic,
proprioceptive, and temporal information. Coupling the training and deployment
infrastructure to these internal choices would require a separate execution
path for every model family. \fluxvla therefore treats the top-level policy as
the model-side integration boundary, behind which architecture-specific
components and algorithms can be implemented independently.

During training, a policy consumes the semantic batch produced by the data
pipeline and returns a primary optimization objective together with optional
diagnostic quantities. During inference, it maps the processed observations
and task specification to an action or action sequence. This compact interface
allows the same execution stack to support autoregressive policies,
continuous-action generative policies, and world--action models despite their
different prediction procedures.

The shared external contract does not require models to expose identical
internal structures. Some policies provide clear boundaries between multimodal
representation learning and action generation, whereas others tightly couple
token generation, dynamics prediction, or iterative action sampling. \fluxvla
allows compatible backbones, projectors, and action modules to be composed
where appropriate, while preserving specialized implementations when the
underlying algorithm requires stronger coupling.

Modular construction therefore guarantees a common integration mechanism
rather than unrestricted interchangeability. Components must still agree on
feature structure, temporal semantics, conditioning information, and action
representation. By making these boundaries explicit, \fluxvla supports model
extension without hiding the mathematical assumptions of individual policy
families.

\paragraph{An optional RL contract.}
Interactive learning extends action prediction with value estimates and the
information needed to recompute rollout probabilities. The RL frontend keeps
the sampled denoising trajectory fixed during learner updates, so rollout and
optimization refer to the same stochastic event. This contract is separate
from the native supervised loss interface and leaves PPO and distributed
orchestration to RLinf. Model and benchmark registrations remain independent,
allowing compatible additions without changing the shared training loop
(\cref{sec:rl-post-training}).

Finally, the native training contract includes a complete and transferable
trained state. The architecture can be reconstructed from the saved experiment specification
and restored from the corresponding model artifact without separately
recovering its original initialization weights. This provides a consistent
handoff from training to evaluation, serving, and physical-robot deployment.
RLinf restoration follows the separate initialization and checkpoint contract
in \cref{sec:rl-post-training}, rather than implicitly inheriting this guarantee.

\subsection{Execution lifecycle and deployment contract}
\label{sec:runner-contract}
\label{sec:runner-operator-contract}

A model defines how actions are predicted, but it does not specify how an
experiment is initialized, distributed, evaluated, or connected to a robot.
Embedding these responsibilities in individual policies would duplicate
infrastructure and tightly couple model development to execution details.
\fluxvla therefore uses runners as the orchestration boundary between model
computation and complete executable workflows.

Runners follow a common lifecycle that covers initialization, repeated
execution, artifact production, and resource release. Training runners
coordinate optimization and checkpointing, evaluation runners manage
environment interaction and metric aggregation, and inference runners maintain
the temporal state required for continuous control. This separation allows the
same policy interface to be used across training, simulation, local inference,
and remote serving without requiring the model to manage workflow-specific
state.

Explicit resource ownership also provides a clean boundary between successive
stages. In particular, training can release its distributed and computational
state before evaluation begins, while the produced model artifact is
transferred to a fresh evaluation process. This design supports a reliable
train--evaluate lifecycle and prevents residual training state from affecting
subsequent evaluation or deployment.

Physical deployment introduces an additional boundary between model-facing
semantics and hardware-specific communication. \fluxvla separates these
responsibilities between the inference runner and the operator. The runner
manages preprocessing, prediction, optional trajectory post-processing,
execution timing, action chunks, and RTC or trajectory-stitching state. The
operator acquires and synchronizes
sensor observations, translates them into the framework representation, and
delivers commands through the robot's communication interface.

\begin{table*}[t]
  \tablestyle
  \caption{Responsibility split at the real-robot boundary. The runner owns
  policy execution and temporal state, while the operator owns hardware-facing
  communication and command transport.}
  \label{tab:runner-operator}
  \small
  \setlength{\tabcolsep}{5pt}
  \begin{tabularx}{\linewidth}{@{}CC@{}}
    \toprule
    Inference runner & Operator \\
    \midrule
    Construct the model or remote client and load run-local artifacts
      & Initialize robot middleware and vendor communication \\
    \addlinespace[3pt]
    Maintain observations, instructions, action chunks, timestamps, RTC
    context, and trajectory-stitching state
      & Acquire and synchronize sensors and maintain frame queues \\
    \addlinespace[3pt]
    Invoke prediction, convert outputs to robot-space actions, and optionally
    post-process continuous trajectories
      & Publish joint, end-effector, gripper, head, or whole-body commands \\
    \addlinespace[3pt]
    Decide when a chunk is replaced, truncated, stitched, or canceled
      & Execute and stop trajectories and report their active state \\
    \addlinespace[3pt]
    Release model-side, serving-client, and temporal state
      & Release communication resources, subscribers, queues, and threads \\
    \bottomrule
  \end{tabularx}
\end{table*}

This division keeps temporal policy logic independent of a particular robot
platform while allowing the same operator abstraction to support different
models and serving modes. Observation, prediction, and communication failures
are propagated explicitly so that the surrounding workflow can stop, retry, or
recover without fabricating valid inputs. Hardware safety mechanisms and
low-level control constraints remain the responsibility of the robot controller
and deployment environment.

\subsection{Extension recipes}
\label{sec:extension-recipes}

The contracts above reduce common integrations to three bounded recipes. Each
recipe begins by choosing the narrowest extension point that preserves the
existing semantics; a new top-level runner or model should not be introduced
when a transform or adapter is sufficient.

\paragraph{Adding a dataset.}
First, convert the source trajectories to the common Parquet-oriented episode
layout when possible, or implement a registered dataset that yields temporally
valid sample dictionaries. Second, define field mappings, observation and
action horizons, dataset/group statistics, and an ordered list of registered
transforms that produces the keys expected by the target model. Third, classify
every final field in the configured collator, stacking tensor fields and
preserving required metadata. Finally, inspect samples at episode boundaries
and verify that normalization followed by denormalization recovers the intended
physical action convention. No change to the training loop is required when
the resulting batch already satisfies the model contract.

\paragraph{Adding a model.}
First, determine whether the integration is a new composition of existing
modules or a new algorithmic owner. Register reusable backbones, projectors, or
heads separately only when their interfaces are meaningful outside one model.
Implement a top-level VLA that builds its children, defines \texttt{forward},
returns a scalar \texttt{loss}, exposes \texttt{predict\_action}, and supplies
an FSDP wrapping policy when distributed training is required. Then define the
model-specific transforms and collator keys, state/action dimensions, temporal
horizon, and checkpoint name mapping. Validate construction, one forward and
backward step, checkpoint reload, and action prediction before attaching the
model to benchmark or robot runners.

\paragraph{Adding a robot.}
First, implement or reuse an operator that synchronizes the necessary camera
and state streams and provides the required command transports. Preserve
source timestamps and define the logical observation names consumed by the
online data pipeline. Second, implement the narrow robot-specific runner
methods that assemble the observation window and execute denormalized actions;
reuse the base preprocessing, prediction, postprocessing, serving, and cleanup
paths. Third, configure state and action dimensions, camera order, control
representation, publish rate, action chunk, execution horizon, statistics key,
and any optional trajectory post-processing settings. Before closed-loop
policy execution, validate observation freshness,
normalization, command signs and units, trajectory cancellation, and external
safety interlocks with a non-learning command sequence.

Together, these recipes illustrate the purpose of stable contracts in
\fluxvla. Extensibility does not mean that every component can be combined
with every other component. It means that incompatibilities are expressed at a
small number of reviewable boundaries, and that a compatible new dataset,
model, execution backend, or robot can reuse the remainder of the system
without copying its training and deployment infrastructure.

%% file: sections/05_experiments.tex
\section{Experimental Evaluation}
\label{sec:experiments}

The preceding chapters describe the mechanisms by which \fluxvla unifies VLA
training and deployment. This chapter asks whether the released system provides
corresponding empirical evidence. For an engineering platform, evaluation must
cover more than the highest task-success number. It should establish that
heterogeneous policy families can execute through the common stack, that
closed-loop performance remains competitive after integration, that runtime
optimizations improve execution without changing the intended computation,
that trajectory post-processing can be integrated with asynchronous chunk
execution at measurable runtime cost, and that the deployment path reaches
physical robots under explicit operational contracts.

The current evidence has three different levels of provenance. First, the
source tree itself provides auditable configurations, component registrations,
tests, evaluators, checkpoint conventions, and robot runners. Second, the
project reports simulation and inference measurements together with six-task
ALOHA, four-task Oli, four-task Rebot, and five-task dual-arm Franka benchmarks.
Third, a companion study reports a quantitative real-robot
result obtained with the platform. We keep these levels
separate throughout the chapter. Values transcribed from project documentation
are labeled \textbf{project-reported}; they are not presented as independently
rerun measurements. Likewise, the existence of a robot operator is treated as
integration coverage, not as evidence of task success.

\subsection{Evaluation}
\label{sec:evaluation-questions}

We evaluate \fluxvla along three dimensions corresponding to its primary system
objectives.

\paragraph{Closed-loop effectiveness.}
We evaluate whether policies integrated through \fluxvla retain reliable
task-level performance on representative simulation benchmarks under the
shared training and evaluation pipeline.
For online post-training, we additionally define matched base-policy versus
RL comparisons to distinguish policy improvement from protocol changes.

\paragraph{Runtime efficiency.}
We report the inference-frequency improvements of the optimized execution
paths, assess numerical consistency for the tested GR00T action-head paths,
and measure the host-side latency added by the trajectory-processing backends.

\paragraph{Deployment capability.}
We assess the extent to which the unified execution stack supports end-to-end
physical deployment, including the combined RTC and trajectory-processing
path, and distinguish quantitative closed-loop results from integration and
demonstration coverage.

\fluxvla is a system framework rather than a new policy-learning algorithm.
Accordingly, the experiments focus on integration coverage, execution
correctness, runtime efficiency, and deployment readiness. Direct comparisons
between policy families would require controlled pretraining and task-data
budgets, optimization budgets, and evaluation protocols beyond the scope of
this work, with architecture treated as the factor being compared.

\subsection{Evaluation protocol}
\label{sec:evaluation-protocol}

Our evaluation covers three complementary aspects of \fluxvla: closed-loop
policy performance in simulation, inference efficiency, and deployment on
physical robots. For each quantitative result, we state the available model,
protocol, and provenance information. Some project-reported summaries omit
trial-level records or complete settings; their limitations are stated rather
than treated as a fully reproducible comparison. Implementations with no
reported task outcomes establish integration coverage only.

\paragraph{Simulation.}
We evaluate closed-loop performance on LIBERO, RoboCasa GR-1, RoboTwin,
and RoboDojo. Task success rate is the primary metric; RoboDojo additionally
reports progress scores to distinguish partial completion from full-task
success. On LIBERO, results
are reported over the commonly used Spatial, Object, Goal, and Long
(\texttt{libero\_10}) suites. The first three emphasize controlled changes in
spatial relations, object identities, and goal specifications, while the Long
suite contains ten longer-horizon tasks drawn from LIBERO-100
~\cite{libero2023}. Under the project-specific RoboCasa GR-1 protocol,
results are grouped into trained-task categories and a generalization set; the
tasks are instantiated in the RoboCasa environment family~\cite{robocasa2024}.
On RoboTwin, we report project-provided Clean and Random success rates
for eight integrated policies, with Average defined as the arithmetic mean
of the two settings. The source summary does not specify trial counts, training budgets, or
per-task uncertainty for these results, so they are not treated as matched
comparisons between policy families. This evaluation is separate from the ten-task
SmolVLA paired RL evaluation below.
On RoboDojo, we follow the project-reported 42-task evaluation for eight
integrated policies. Each of 30 standalone
tasks uses 50 trials, and each of 12 generalization tasks uses 25
standard-layout and 25 randomized-layout trials, totaling 2,100 trials per
model. Results include Average, Gen-Std, Gen-Rand, Precision, Long-Horizon,
Memory, and Open, with progress score and success rate reported separately.
RoboDojo averages are retained from the source; we do not substitute an
unweighted mean of the displayed group-level values or infer uncertainty
from the rounded summary statistics.
Success rates are aggregated
from task-level rollouts under the reported protocol for each integration.
For GPT-6, resource constraints limit evaluation to five episodes per task,
giving 50 episodes for each ten-task suite and 200 episodes in total.
These limited-sample estimates are reported separately from the
checkpoint-based results rather than assumed to use a matched evaluation
budget.

\paragraph{Reading the overview performance plot.}
\Cref{fig:system-overview} plots the reported average success percentages
on a shared linear 0--100\% axis, not relative gains or scores normalized
to a baseline. RoboTwin uses the Clean/Random mean; RoboDojo uses full-task
success rather than progress. Lines connect categorical integration entries,
not a continuous trend; stars and bold labels mark each benchmark's maximum.
The Cosmos3 label groups Nano results on LIBERO, RoboCasa, and RoboDojo with
the unspecified Cosmos3 variant on RoboTwin, without implying identical
variants or checkpoints. Tasks, protocols, and training budgets differ,
so the plot neither establishes a controlled architecture ranking nor
supports direct comparisons of benchmark difficulty. Full results appear in
\cref{tab:libero-results,tab:robocasa-results,tab:robotwin-results,tab:robodojo-results}.

\paragraph{Inference efficiency.}
We report project-documented baseline and accelerated model frequencies for
each listed model/device case. A controlled rerun should hold the checkpoint,
hardware, precision, input shapes, and denoising schedule fixed; the available
timing summaries do not fully document these settings
(\cref{tab:runtime-results}). Model-path frequency is distinct from end-to-end
robot control frequency, which also includes sensing, preprocessing,
communication, and command execution. Numerical comparisons are reported
separately for the tested random-weight GR00T heads
(\cref{tab:runtime-parity}), not as evidence of full-model or task-level parity
for every accelerated path.

\paragraph{Real-robot evaluation.}
The ALOHA benchmark comprises clothes folding, red-towel folding, and four
pick-and-place tasks using a bear, a white racing car, a purple caterpillar
toy, and a brown bird toy. The folding tasks use 20 and 30 trials,
respectively, and each pick-and-place task uses 10 trials. For all seven
reported policies, overall success pools successful episodes over all 90
trials rather than averaging the six task-level rates.
The Oli benchmark covers candy picking, clearing four tabletop objects into
a basket, four-stage box transport, and basket-and-toy picking.
The author-specified trial budgets are 120, 60, 60, and 60, respectively.
GR00T N1.5 candy picking records 49 successes in 120 trials, yielding
40.83\% after rounding to two decimal places. For $\pi_{0.5}$ box transport,
37/60 trials complete stages 1--3 and 20/60 complete all four stages, yielding
61.67\% and 33.33\%, respectively. Cosmos3 basket-and-toy picking succeeds
in 7/60 trials (11.67\%). Other percentages retain the reported precision;
no pooled Oli success rate is computed. For box transport,
completion of stages 1--3 is an intermediate milestone, whereas completion
of all four stages measures full task success. Missing results are not
treated as failures.
Rebot evaluates block transfer, box opening and car placement, flower
insertion, and black-block stacking. Each policy uses 20 trials per task,
and the overall success rate pools all 80 trials across the four tasks.
The dual-arm Franka benchmark evaluates shuttlecock-tube handover, bag loading
with left-arm bag manipulation and right-arm object placement, alternating
chick-toy picking, coordinated drawer opening and object placement, and block
stacking. Each task uses 20 trials, except SmolVLA's last three tasks, which
use 10 trials each. Overall success therefore pools 100 trials per policy,
or 70 for SmolVLA. Both GR00T configurations freeze the VLM; the reported
DiT4DiT, FastWAM, and Cosmos3 configurations use RTC.
The companion ARM study follows a separate task and evaluation protocol, so its
results are reported independently. Configurations and demonstrations without
trial outcomes establish deployment compatibility rather than success rates.

For reproducibility, benchmark assets, evaluation settings, random seeds,
action horizons, and checkpoint-selection rules should remain fixed within
each comparison. We provide the associated configurations, checkpoints, and
evaluation entry points whenever available, enabling the reported experiments
to be reproduced without relying on implementation-specific assumptions.

\subsection{Simulation benchmark results}
\label{sec:simulation-results}

\Cref{tab:libero-results} reports fourteen integrations: thirteen
checkpoint-based policies spanning compact VLA, pretrained flow-matching,
modular VLM/action-head, and world--action architectures, plus an inference-only
GPT-6 API policy. Average success ranges from 37.50\% for GPT-6 to 98.65\% for
DiT4DiT; eleven integrations exceed 95\%, and the highest Long-suite result is
98.0\% for Cosmos3-Nano. These results demonstrate shared evaluation across heterogeneous
policy implementations, not a controlled leaderboard: pretraining,
optimization, checkpoint selection, control interfaces, and evaluation budgets
are not matched across rows.

\input{tables/simulation_benchmarks}

The suite-level pattern is also informative. SmolVLA's largest reduction is on
the Long suite, while several larger or generative integrations retain strong
performance there. By contrast, the Object suite is at or above 98.8 for eight
reported integrations. This suggests that the evaluation path is not saturated
uniformly across task types, although identifying the cause would require
matched training budgets and per-task outcomes. The table reports uncertainty
only for the three Long entries that include it in the source record; the absence
of an interval elsewhere must not be read as zero variance.

\Cref{tab:robocasa-results} extends the evaluation beyond LIBERO and exposes a
more challenging regime. Across the eight reported integrations, average
success ranges from 8.75\% for SmolVLA to 57.25\% for DiT4DiT. GR00T N1.5,
trained with 30 demonstrations per task, reports 44.30\%, while GR00T N1.7,
trained on the full dataset, reports 46.42\%. FastWAM reaches 49.92\% overall
and 50.00\% on the Generalization group. The full-data $\pi_0$,
$\pi_{0.5}$, and DiT4DiT integrations report 51.00\%, 51.42\%, and 57.25\%,
respectively. These values are not a controlled architecture comparison
because the data budgets, model initialization, and optimization procedures
are not matched, and the source does not report per-task uncertainty.

\Cref{tab:robotwin-results} adds RoboTwin Clean and Random results for eight
\fluxvla integrations. FastWAM has the highest success rate among these
integrations in both settings, at 90.48\% and 90.18\%, respectively,
with a two-setting average of 90.33\%.
GR00T N1.5 reports 89.42\% on Clean and 88.56\% on Random, while GR00T N1.7
reports 88.56\% and 89.50\%. Their averages are 88.99\% and 89.03\%,
respectively. DiT4DiT and Cosmos3 achieve 77.22\%/81.66\%
and 78.12\%/75.04\%, respectively. The $\pi_0$ integration reports
67.86\%/67.90\%, and $\pi_{0.5}$ reports 81.46\%/84.02\%.
The corresponding averages are 79.44\% for DiT4DiT, 76.58\% for Cosmos3,
67.88\% for $\pi_0$, and 82.74\% for $\pi_{0.5}$.

\input{tables/robotwin_benchmarks}

\Cref{tab:robodojo-results} reports a more challenging evaluation profile
across eight integrated policies on RoboDojo. The reported average progress
score / success rate pairs are 3.03 / 0.93\% for Cosmos3-Nano,
3.08 / 1.45\% for GR00T N1.7, 4.84 / 2.45\% for DiT4DiT,
5.29 / 2.98\% for SmolVLA, 5.29 / 2.97\% for FastWAM,
5.50 / 3.17\% for GR00T N1.5, 5.96 / 3.37\% for $\pi_0$,
and 13.61 / 8.83\% for $\pi_{0.5}$. Among these integrations, $\pi_{0.5}$ leads both average
metrics as well as the Gen-Std, Gen-Rand, Precision, Long-Horizon, and Open
metrics. GR00T N1.5 leads Memory with a progress score of 5.48 and a success
rate of 4.33\%.

Generalization is particularly sensitive to layout changes in these reported
results. For $\pi_{0.5}$, success decreases from 16.00\% on Gen-Std to
2.33\% on Gen-Rand, while progress decreases from 22.65 to 6.25. The other
seven models report randomized-layout success rates of at most 0.67\%.
Even the highest Long-Horizon result, 28.04 progress and 18.75\% success
for $\pi_{0.5}$, leaves substantial room for improvement. RoboDojo therefore
complements RoboTwin's Clean/Random comparison by exposing low full-task
success and differences between partial progress and completed tasks.
These results characterize the evaluated checkpoints and protocols; they
do not establish a controlled architecture ranking or permit direct
cross-benchmark comparisons of difficulty.

\input{tables/robodojo_benchmarks}

These results demonstrate strong closed-loop LIBERO performance for multiple
checkpoint-based policy families and extend the shared \fluxvla evaluation
pipeline to an API-hosted controller. They also validate the RoboCasa GR-1
evaluation path across eight model and data configurations and add RoboTwin
results across eight policy integrations. RoboDojo further extends the
evaluation to eight integrations across 42 tasks with both progress and
success metrics. We do not claim exact
reproduction of every upstream implementation, as such comparisons require
matched checkpoints, preprocessing, simulator versions, random seeds, and
evaluation protocols.

\subsection{Policy improvement through reinforcement learning}
\label{sec:rl-results}

We evaluated whether the online RL integration described in
\cref{sec:rl-post-training} improves a supervised VLA policy under paired
simulator evaluation. FluxVLA provides the model front end, benchmark-specific
observation and action adapters, and flow-matching policy interface, whereas
RLinf~\cite{yu2026rlinf,rlinfvla2025} provides synchronous environment
rollout, PPO with generalized advantage
estimation, FSDP2 training, actor--rollout weight synchronization, and
checkpoint management. During training, one randomly selected denoising
transition is sampled with Flow-SDE~\cite{pirl2025}, and the learner recomputes its probability
from the same stored latent transition. Evaluation instead uses the policy's
native deterministic ODE inference procedure.

\input{tables/rl_post_training}

\paragraph{Task-wise reinforcement learning.}
For each benchmark, all tasks were initialized from the same SFT checkpoint.
We independently optimized a separate policy for each task using PPO.

\paragraph{Paired evaluation.}
For each task, we evaluated the SFT and RL policies using the same simulator
settings, observation preprocessing, normalization statistics, action and
execution horizons, and denoising schedule. Evaluation used native ODE
inference with matched episode initializations and inference-noise seeds.
We averaged per-task success rates across tasks and computed 95\% confidence
intervals for paired improvements using 100,000 bootstrap resamples within
each task. All reported runs use training seed 1234; these intervals quantify
episode-sampling uncertainty for the evaluated checkpoints, not variability
across independent training runs. These paired RL experiments are separate
from the multi-policy simulation tables and should not be compared to their
SmolVLA baselines without matching checkpoints, task sets, and protocols.

\paragraph{Current evidence.}
The integration provides regression tests for native inference parity,
rollout--learner probability consistency, gradient flow, parameter freezing,
and checkpoint restoration. These establish interface-level checks rather
than benchmark improvements. Separately,
\cref{tab:rl-post-training-results} reports paired success-rate measurements
for SmolVLA on LIBERO-10 and ten RoboTwin clean tasks, as detailed below.

\paragraph{RoboTwin.}
We evaluated SmolVLA on ten representative RoboTwin clean tasks spanning
object manipulation, tool use, bimanual coordination, handover, and stacking.
Each task used 968 training seeds, 32 validation seeds, and 100 fixed test
seeds. Starting from the same SFT checkpoint, we independently trained a
task-specific policy for 40 PPO outer steps, collecting 128 episodes per step.
The evaluation therefore contained 1,000 paired test episodes across ten
tasks. The macro-average success rate increased from 45.9\% for SFT to
57.9\% at step 20 and 61.3\% at step 40. The step-40 policy improved the SFT
baseline by 15.4 percentage points (paired bootstrap 95\% CI, 12.2--18.6),
with 231 rescued and 77 regressed episodes. Nine of the ten task-specific
policies improved over their SFT initialization.

\paragraph{LIBERO-10.}
We independently fine-tuned SmolVLA on all ten LIBERO-10 tasks using PPO
and evaluated each policy on 50 matched trials per task.
The mean success rate increased from 66.2\% for SFT to 71.0\%, 76.2\%,
and 74.2\% after 20, 40, and 60 PPO steps, respectively.
The highest observed mean was achieved at step 40, improving over SFT
by 10.0 percentage points, with gains on nine tasks and unchanged
performance on one. Further training to step 60 reduced the mean success
rate, showing that improvement was not monotonic with training duration.

\subsection{Inference acceleration and numerical alignment}
\label{sec:runtime-results}

The runtime evaluation compares the ordinary model path with inference-specific
implementations that replace eligible operations with Triton fused kernels,
CUDA Graph replay, and custom CUDA operators. The optimized GR00T path uses an
inference VLM backbone and flow-matching head, while the optimized
$\pi_{0.5}$ path captures a larger portion of the vision--language--action
pipeline. These are model-specific optimizations behind the same
\texttt{predict\_action} boundary; they are not a universal compiler pass over
arbitrary registered models.

\input{tables/runtime_benchmarks}

Across the eight reported device/model cases in
\cref{tab:runtime-results}, acceleration raises model frequency by
$2.31\times$ to $9.64\times$. On A100, GR00T increases from 5.96 to 32.6 Hz and
$\pi_{0.5}$ from 2.2 to 21.2 Hz. On RTX 5090, the optimized paths reach 42.6 Hz
for GR00T, 47.6 Hz for GR00T with Training-time RTC, and 31.6 Hz for
$\pi_{0.5}$. The edge
result is also practically relevant: on AGX Orin 64GB, GR00T rises from 3.2 to
7.4 Hz and $\pi_{0.5}$ from 1.4 to 4.4 Hz. The magnitude varies across devices
and models because the fraction of work captured or fused, launch overhead,
sequence shapes, and accelerator characteristics differ.

The 4090 $\pi_{0.5}$-RTC source row, which uses the Training-time RTC route,
lists 3.4 and 19.6 Hz. Their ratio is
$5.76\times$, which is the value reported here; this corrects the inconsistent
$6.66\times$ ratio printed in the frozen acceleration document. More broadly,
the frequencies require a standardized rerun before publication-quality
hardware comparison. The current documentation does not completely specify
warm-up iterations, measurement repetitions, synchronization points, batch and
sequence shapes, denoising steps, clock/power settings, or whether image
preprocessing is inside the timed region.

Acceleration is useful only if it preserves the intended policy computation.
The released GR00T test copies matched random weights into the baseline and
accelerated heads, holds inputs and sampling seed fixed, and compares predicted
actions. Cosine similarities exceed 0.99999 for both the plain and
Training-time RTC prefix-conditioning paths, with maximum absolute differences
of approximately 0.02. These results
are consistent with small changes from bf16 kernels, fusion, and operation
ordering. They validate numerical alignment of the tested head-level paths,
but not full checkpoint equivalence or task-level parity. A stronger final
protocol should add representative trained checkpoints, multiple observations
and seeds, action-error distributions after denormalization, and closed-loop
success with and without acceleration.

These results characterize runtime efficiency in two parts. First, specialized
backends provide substantial model-path speedups across datacenter, desktop,
and edge GPUs. Second, the tested accelerated GR00T head remains closely
aligned with the baseline computation. End-to-end control benefit remains a
separate measurement: an inference path running at 40 Hz does not imply a
40-Hz robot loop if cameras, network transfer, or command execution are slower.
The serving profiler already separates serialization, round trip, server
inference, network residual, deserialization, payload size, and total time; a
future quantitative table should aggregate those measurements under controlled
local and remote network conditions.

\subsection{Trajectory post-processing behavior}
\label{sec:trajectory-postprocessing-results}

We benchmark the host-side latency of the joint MPC and Ruckig backends
independently of policy inference. Both implementations process each selected
joint separately; this experiment does not evaluate coupled-joint or Cartesian
post-processing.

\begin{table}[t]
  \tablestyle
  \caption{Project-reported trajectory post-processing latency for a 12-DoF,
  50-step synthetic trajectory on an Intel Xeon Platinum 8336C CPU. Values
  include warmed solver execution but exclude policy inference and robot I/O.
  Bold denotes the lower latency.}
  \label{tab:trajectory-postprocessing-runtime}
  \small
  \setlength{\tabcolsep}{7pt}
  \begin{tabular}{@{}ccc@{}}
    \toprule
    Backend & Mode & Mean latency (ms) \\
    \midrule
    Joint MPC & Tracking & 4.90 \\
    Ruckig filter & Tracking & \textbf{1.17} \\
    \bottomrule
  \end{tabular}
\end{table}

The measured tracking latency in
\cref{tab:trajectory-postprocessing-runtime} is small relative to many
large-model inference calls, although it is not negligible for high-rate
control. Ruckig is approximately $4.19\times$ faster in this configuration,
showing that backend choice changes the CPU overhead added before command
execution. These measurements are specific to the reported processor, horizon,
and degree-of-freedom count and should not be interpreted as a general hardware
ranking. With bounds chosen for the robot, the joint-separable formulation
provides a practical precision--smoothness trade-off: looser bounds track the
policy reference more closely, whereas tighter bounds reduce higher-order
motion at the cost of tracking error or lag. This result characterizes the two
provided backends rather than the broader post-processing interface, which can
host other trajectory-processing methods.

\begin{figure*}[t]
  \centering
  \includegraphics[width=0.96\textwidth]{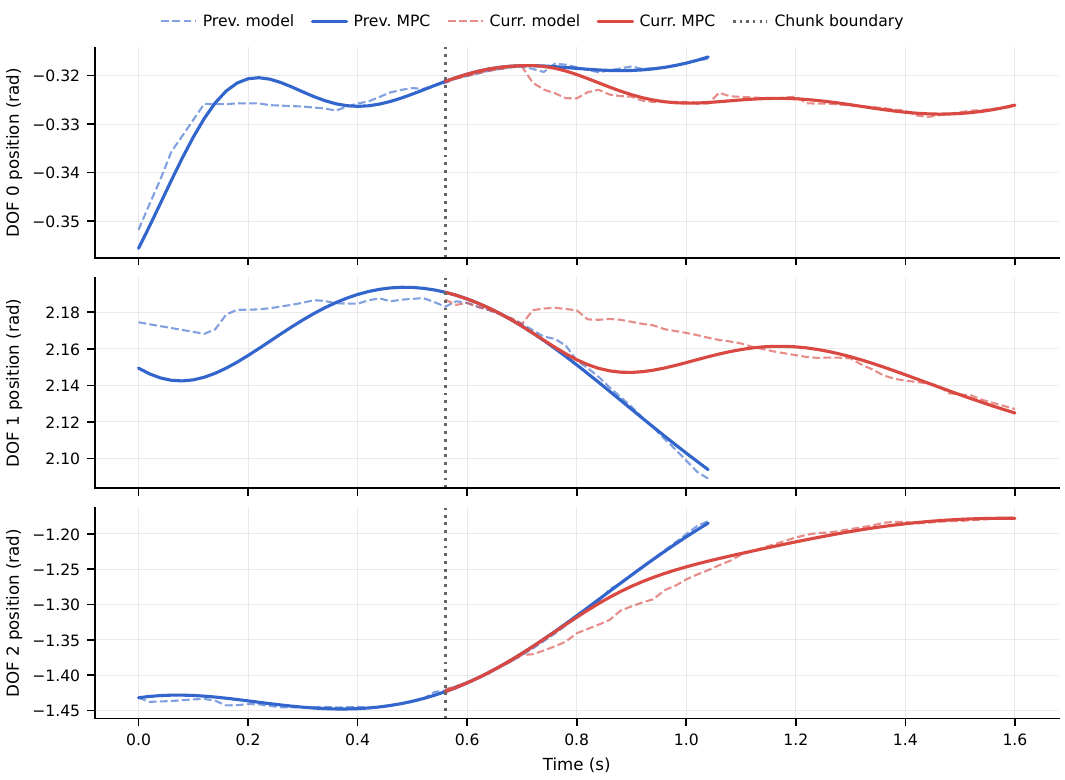}
  \caption{Qualitative real-robot runtime trace of asynchronous Training-time
  RTC followed by joint-MPC trajectory post-processing. Dashed and solid curves
  denote model outputs and MPC-processed trajectories, respectively; blue and
  red indicate the previous and current chunks, and the dotted line marks the
  chunk boundary. Across three representative joints, the current MPC
  trajectory connects to the active previous trajectory at the handoff.}
  \label{fig:trajectory-postprocessing-stitch-runtime}
\end{figure*}

Together, the latency measurement and the runtime trace in
\cref{fig:trajectory-postprocessing-stitch-runtime} provide complementary
runtime and deployment evidence: trajectory post-processing adds
millisecond-scale host overhead, and a Training-time RTC output chunk is
post-processed to connect to the active processed trajectory at an asynchronous
handoff. The trace is qualitative and does not isolate Training-time RTC from
post-processing or establish an effect on task-level performance.

\subsection{Real-robot benchmark results}
\label{sec:real-robot-results}

\Cref{tab:aloha-benchmark-results} reports seven policy configurations, all
with complete six-task results. $\pi_{0.5}$ has the highest pooled success rate at
72.22\% (65/90), followed by GR00T N1.7 at 62.22\%
(56/90) and FastWAM at 61.11\% (55/90). DiT4DiT with RTC, GR00T N1.5,
Cosmos3, and SmolVLA achieve 54.44\%, 47.78\%, 41.11\%, and 8.89\%,
respectively. Performance varies by task: GR00T N1.7 has the highest
red-towel-folding rate (20/30), while FastWAM succeeds in all ten bear
pick-and-place trials. These results extend physical deployment evidence
across VLA and world--action policies, but characterize the reported
configurations rather than isolate architecture, pretraining, or RTC effects.

\input{tables/real_robot_benchmarks}

\Cref{tab:oli-benchmark-results} extends the comparison to five policy
configurations on Oli, with DiT4DiT reported only on the two stationary
tasks. $\pi_{0.5}$ has the highest reported candy-picking and tabletop-clearing
rates (62.5\% and 45.0\%). On box transport, $\pi_{0.5}$ and FastWAM report
33.33\% and 31.7\% full-task success, respectively, compared with 6.7\% for
GR00T N1.5 and Cosmos3, despite much closer stage-1--3 completion rates.
FastWAM's 48.3\% basket-and-toy result uses an intermediate starting point
and therefore does not establish completion of the full navigation sequence.
These results characterize the recorded
configurations and starting conditions, not a controlled architecture ranking.

\Cref{tab:rebot-benchmark-results} adds four Rebot manipulation tasks across
seven policy configurations. $\pi_{0.5}$ has the highest reported success
rates on block transfer, box opening and car placement, and flower insertion
(95.00\%, 85.00\%, and 75.00\%), and the highest pooled rate (77.50\%).
GR00T N1.7 has the highest stacking rate at 70.00\%, compared with
DiT4DiT's 55.00\%. These results provide additional
cross-platform deployment evidence, without isolating architecture or
training effects.

\Cref{tab:franka-benchmark-results} further evaluates bimanual coordination
on Franka across seven policy configurations. GR00T N1.7 has the highest
pooled success rate at 88.00\%, leading bag loading, alternating toy picking,
and drawer loading at 95.00\%, 95.00\%, and 90.00\%, respectively.
$\pi_{0.5}$ achieves the highest tube-handover and block-stacking rates
(85.00\% each), with 79.00\% pooled success. The results characterize the
reported training and execution configurations rather than isolate the
effects of model architecture, VLM freezing, or RTC. SmolVLA's smaller trial
budget also changes the task weighting of its pooled score.

\Cref{tab:arm-real-robot-results} provides complementary evidence from the
companion ARM study. On a separate eight-stage ALOHA towel-folding task, the
reported success rate rises from
62.1\% for behavior cloning to 78.5\% for SARM-based RA-BC and 99.4\% for
ARM-based AW-BC, alongside higher episode throughput and folding precision.
The result demonstrates that \fluxvla can participate in a closed loop spanning
real-robot data, reward modeling, reweighted policy training, and deployment.
Because the study does not state the number of policy-evaluation trials, this
report preserves the published point estimates but does not derive confidence
intervals or claim statistical significance.

A common real-robot protocol should evaluate a small representative subset
rather than every available platform. For each policy--robot pair, it should
freeze the checkpoint, camera layout, control representation, normalization
statistics, action horizon, execution horizon, inference location, and control
rate. Each task should declare its initial-state distribution, timeout, success
predicate, number of independent trials, intervention rule, and treatment of
communication or safety failures. Reporting should include per-task success
with confidence intervals, end-to-end latency, intervention rate, completion
time, and a failure taxonomy. These records are necessary to distinguish model
errors from sensing, transport, controller, and operator failures.

\subsection{Findings and remaining controlled studies}
\label{sec:evaluation-findings}

The current evidence supports three conclusions. First, across multiple
integrated policy families, the released configurations obtain strong
closed-loop results on LIBERO and execute through RoboCasa GR-1 and RoboTwin,
with eight policy integrations evaluated in RoboTwin's Clean and Random
settings. RoboDojo adds a 42-task evaluation of eight integrations,
exposing gaps in full-task completion and randomized-layout generalization
through complementary progress and success metrics. Second, model-specific accelerated backends substantially
increase reported inference frequency with close numerical alignment in the
available GR00T parity test; the trajectory benchmark additionally quantifies
the host-side latency of joint MPC and jerk-limited filtering. Third, the
six-task ALOHA, four-task Rebot, and five-task Franka benchmarks cover multiple
VLA and world--action policy configurations, while the four-task Oli benchmark provides stage-aware
stationary and mobile manipulation results across additional policy families.
The ARM study
provides complementary evidence for offline policy improvement.
The optional RLinf frontend adds an online improvement path, with paired
SmolVLA gains of 10.0 percentage points on LIBERO-10 and 15.4 percentage
points on ten RoboTwin clean tasks reported in
\cref{tab:rl-post-training-results}.
The runtime trace additionally demonstrates that
Training-time RTC and trajectory post-processing can execute in sequence.

Several claims remain deliberately open. The current snapshot does not provide
a uniform upstream-versus-FluxVLA reproduction study, a controlled comparison
of all LIBERO models, multi-seed uncertainty for most table entries, a
single-/multi-node training scaling sweep, task-level parity for accelerated
inference, a controlled comparison of RTC routes under injected latency, a local-versus-remote
closed-loop comparison, or a common multi-robot trial benchmark. These are the
highest-value additions for a subsequent report revision because each isolates
a platform claim rather than adding another uncalibrated model score.

A suitable controlled-study order is: (1) reproduce one autoregressive and one
flow-matching policy against their upstream evaluation paths; (2) benchmark
training throughput and peak memory as device count increases; (3) measure
baseline, individual kernel groups, CUDA Graphs, and their combination under a
single latency harness; (4) compare an unconditioned baseline, Training-time
RTC, original Test-time RTC with VJP guidance, and the direct approximation
across controlled inference delays and execution horizons; and (5) run a shared
pick-and-place protocol on two
embodiments with both local and remote inference. This sequence would turn the
current breadth and project-reported measurements into causal evidence about
reproduction, scaling, runtime optimization, temporal coordination, and
cross-platform deployment.

%% file: tables/simulation_benchmarks.tex
\begin{table*}[t]
  \tablestyle
  \caption{LIBERO closed-loop success rates (\%) for integrated policy
  checkpoints and the inference-only GPT-6 API policy. Uncertainty is shown
  only where reported; $^{\dagger}$ marks the resource-limited evaluation.
  Bold denotes the highest value in each metric column, including ties.}
  \label{tab:libero-results}
  \footnotesize
  \setlength{\tabcolsep}{4.6pt}
  \begin{tabular}{@{}cccccc@{}}
    \toprule
    FluxVLA integration & Spatial & Object & Goal & Long & Average \\
    \midrule
    GPT-6$^{\dagger}$       & 32.0 & 66.0 & 38.0 & 14.0 & 37.50 \\
    SmolVLA                 & 86.2 & 92.4 & 91.4 & 68.8 & 84.70 \\
    Cosmos3-Edge            & 95.6 & 95.6 & 91.6 & 94.8 & 94.40 \\
    GR00T N1.5              & 97.4 & 96.2 & 94.6 & $93.0 \pm 1.5$ & 95.30 \\
    GR00T N1.7              & 96.5 & 97.6 & 97.6 & $92.0 \pm 1.0$ & 95.93 \\
    Qwen3VL-0.6B + GR00T N1.5 & 96.0 & 99.4 & 95.2 & 94.2 & 96.20 \\
    DreamZero               & 98.2 & 98.8 & 93.2 & 94.8 & 96.25 \\
    $\pi_0$                 & 98.6 & 98.8 & 96.8 & 93.2 & 96.85 \\
    Cosmos3-Nano            & 96.0 & \textbf{99.6} & 94.0 & \textbf{98.0} & 96.90 \\
    FastWAM                 & 96.6 & 99.4 & 97.6 & 96.2 & 97.45 \\
    $\pi_{0.5}$             & 98.6 & \textbf{99.6} & 98.0 & $95.6 \pm 1.0$ & 97.95 \\
    FastWAM-IDM             & \textbf{99.8} & 98.0 & 98.4 & 96.2 & 98.10 \\
    FastWAM-Joint           & 99.2 & 98.8 & \textbf{99.6} & 95.8 & 98.35 \\
    DiT4DiT                 & 99.6 & 99.2 & \textbf{99.6} & 96.2 & \textbf{98.65} \\
    \bottomrule
  \end{tabular}

  \par\nobreak\vspace{4pt}
  \noindent\parbox{0.97\textwidth}{\footnotesize\raggedright
  \textbf{Evaluation scope.} The rows differ in architecture, pretraining,
  optimization, and evaluation budget, and are not a controlled algorithmic
  comparison. ``Long'' denotes \texttt{libero\_10}.
  \textbf{$^{\dagger}$GPT-6 evaluation budget.} Due to resource constraints,
  only \textbf{5 episodes per task} are evaluated: 50 per suite and 200 in
  total. Successful episodes are 16/50 (Spatial), 33/50 (Object), 19/50 (Goal),
  and 7/50 (Long); the average is the mean of the four suite success rates.
  Other rows retain the previously reported checkpoint results.}
\end{table*}

\begin{table*}[t]
  \tablestyle
  \caption{Project-reported RoboCasa GR-1 closed-loop success rates (\%).
  Model configurations and training-data budgets differ; the results are
  not a controlled model comparison. Bold denotes the highest value in each metric column.}
  \label{tab:robocasa-results}
  \small
  \setlength{\tabcolsep}{4.5pt}
  \begin{tabular}{@{}ccccccc@{}}
    \toprule
    Integration & Training data & Cabinet & Drawer & Microwave
      & Generalization & Average \\
    \midrule
    SmolVLA
      & 24 tasks, full data
      & 12.50 & 10.00 & 17.50 & 7.22 & 8.75 \\
    Cosmos3-Nano
      & 24 tasks, full data
      & 46.00 & 39.00 & 40.00 & 41.67 & 41.67 \\
    GR00T N1.5
      & 24 tasks, 30 demos/task
      & 22.7 & 35.7 & 32.5 & 48.9 & 44.30 \\
    GR00T N1.7
      & 24 tasks, full data
      & 53.00 & 51.00 & 47.00 & 45.11 & 46.42 \\
    FastWAM
      & 24 tasks, full data
      & 55.00 & 47.00 & 47.00 & 50.00 & 49.92 \\
    $\pi_0$
      & 24 tasks, full data
      & 60.0 & \textbf{56.0} & 48.0 & 49.33 & 51.00 \\
    $\pi_{0.5}$
      & 24 tasks, full data
      & 60.0 & 51.0 & 52.0 & 50.44 & 51.42 \\
    DiT4DiT
      & 24 tasks, full data
      & \textbf{63.0} & 52.0 & \textbf{59.0} & \textbf{57.0} & \textbf{57.25} \\
    \bottomrule
  \end{tabular}

  \par\nobreak\vspace{4pt}
  \noindent\parbox{0.97\textwidth}{\footnotesize\raggedright
  Cabinet, Drawer, and Microwave each aggregate two trained manipulation
  tasks; Generalization aggregates the remaining 18 post-training novel tasks.
  Overall averages are project-reported across all 24 tasks; the source does
  not provide per-task uncertainty. The stated budget is 50 trials per task,
  but some displayed percentages do not reconcile with integer success
  counts under that budget. Raw counts and aggregation details are needed
  before inferring counts or uncertainty from these summaries.}
\end{table*}

%% file: tables/robotwin_benchmarks.tex
\begin{table*}[t]
  \tablestyle
  \caption{Project-reported RoboTwin success rates (\%) under Clean and
  Random settings for eight FluxVLA integrations. Average is the arithmetic
  mean of Clean and Random. Bold denotes the highest value in each metric column.}
  \label{tab:robotwin-results}
  \small
  \setlength{\tabcolsep}{5pt}
  \begin{tabularx}{0.85\textwidth}{@{}YCCC@{}}
    \toprule
    FluxVLA integration & Clean & Random & Average \\
    \midrule
    $\pi_0$ & 67.86 & 67.90 & 67.88 \\
    Cosmos3 & 78.12 & 75.04 & 76.58 \\
    DiT4DiT & 77.22 & 81.66 & 79.44 \\
    $\pi_{0.5}$ & 81.46 & 84.02 & 82.74 \\
    SmolVLA & 87.02 & 85.88 & 86.45 \\
    GR00T N1.5 & 89.42 & 88.56 & 88.99 \\
    GR00T N1.7 & 88.56 & 89.50 & 89.03 \\
    FastWAM & \textbf{90.48} & \textbf{90.18} & \textbf{90.33} \\
    \bottomrule
  \end{tabularx}

  \par\nobreak\vspace{4pt}
  \noindent\parbox{0.97\textwidth}{\footnotesize\raggedright
  The source summary does not specify trial counts, training budgets,
  checkpoint-selection rules, or per-task uncertainty. These results
  therefore characterize the reported integrations rather than establish
  a controlled comparison between policy families. The paired SmolVLA RL results on ten RoboTwin
  clean tasks are reported separately in
  \cref{tab:rl-post-training-results}.}
\end{table*}

%% file: tables/robodojo_benchmarks.tex
\begin{table*}[t]
  \tablestyle
  \caption{Project-reported RoboDojo results for eight FluxVLA integrations
  across 42 tasks.
  Each cell reports progress score / success rate (\%). Rows follow the
  source's ascending average progress score. Bold denotes the highest
  value for each metric within each column, including ties.}
\label{tab:robodojo-results}
\footnotesize
\setlength{\tabcolsep}{2pt}
\begin{tabularx}{\textwidth}{@{}YCCCCCCC@{}}
  \toprule
  Integration & Average & Gen-Std & Gen-Rand & Precision
    & Long-Horizon & Memory & Open \\
  \midrule
  Cosmos3-Nano & 3.03 / 0.93 & 3.95 / 1.33 & 1.20 / 0.67 & 2.29 / 0.00 & 8.74 / 2.75 & 1.23 / 0.67 & 0.30 / 0.25 \\
  GR00T N1.7 & 3.08 / 1.45 & 6.52 / 4.33 & 0.85 / 0.33 & 2.11 / 0.50 & 6.19 / 1.50 & 3.13 / 2.67 & 0.30 / 0.25 \\
  DiT4DiT & 4.84 / 2.45 & 5.67 / 3.33 & 0.83 / 0.00 & 4.49 / 0.00 & 12.17 / 6.50 & 2.45 / 2.33 & 1.83 / 1.75 \\
  SmolVLA & 5.29 / 2.98 & 8.15 / 5.33 & 0.30 / 0.00 & 8.70 / 4.50 & 8.89 / 4.75 & 3.97 / 2.33 & 0.67 / 0.67 \\
  FastWAM & 5.29 / 2.97 & 7.07 / 3.67 & 0.52 / 0.33 & 3.96 / 0.25 & 14.99 / 9.50 & 2.95 / 2.33 & 0.78 / 0.75 \\
  GR00T N1.5 & 5.50 / 3.17 & 6.48 / 3.33 & 1.62 / 0.67 & 2.38 / 0.00 & 15.36 / 9.25 & \textbf{5.48} / \textbf{4.33} & 0.25 / 0.25 \\
  $\pi_0$ & 5.96 / 3.37 & 13.05 / 9.67 & 0.78 / 0.00 & 7.62 / 4.00 & 11.76 / 6.00 & 3.52 / 2.00 & 0.00 / 0.00 \\
  $\pi_{0.5}$ & \textbf{13.61} / \textbf{8.83} & \textbf{22.65} / \textbf{16.00} & \textbf{6.25} / \textbf{2.33} & \textbf{18.28} / \textbf{10.25} & \textbf{28.04} / \textbf{18.75} & 5.10 / 4.00 & \textbf{2.20} / \textbf{2.00} \\
  \bottomrule
\end{tabularx}

  \par\nobreak\vspace{4pt}
  \noindent\parbox{0.97\textwidth}{\footnotesize\raggedright
  Each of 30 standalone tasks uses 50 trials; each of 12 generalization
  tasks uses 25 standard-layout (Gen-Std) and 25 randomized-layout
  (Gen-Rand) trials, totaling 2,100 trials per model.
  Average values are preserved as reported in the source rather than
  recomputed as an unweighted mean of the six displayed group columns.
  Progress scores are distinct from full-task success rates.
  \textbf{Data provenance.} Results are recorded in the project README
  (\url{https://github.com/FluxVLA/FluxVLA}, revision
  \texttt{35ed1adebc79}, accessed October 16, 2026).
  The source summary does not provide per-task uncertainty or establish
  matched training budgets across models.}
\end{table*}

%% file: tables/rl_post_training.tex
\begin{table*}[t]
  \tablestyle
  \caption{Paired evaluation of SmolVLA before and after PPO post-training.
  Each task was independently initialized from the same SFT checkpoint
  for its benchmark. After RL reports the step-40 checkpoint.
  Success rates are task macro-averages in percent; $\Delta$ denotes
  the improvement in percentage points.}
  \label{tab:rl-post-training-results}
  \small
  \setlength{\tabcolsep}{6pt}
  \begin{tabularx}{\textwidth}{@{}CCCCCC@{}}
    \toprule
    Benchmark & Task scope & Policy & Base (SFT) & After RL & $\Delta$ (pp) \\
    \midrule
    LIBERO & Long (10 tasks) & SmolVLA
      & 66.2 & \textbf{76.2} & +10.0 \\
    \addlinespace[3pt]
    RoboTwin & Clean (10 tasks) & SmolVLA
      & 45.9 & \textbf{61.3} & +15.4 \\
    \bottomrule
  \end{tabularx}
  \par\nobreak\vspace{4pt}
  \noindent\parbox{0.97\textwidth}{\footnotesize\raggedright
    Each SFT--RL pair used matched preprocessing, episode initializations,
    inference-noise seeds, action horizons, and native ODE inference.
    Evaluation comprised 50 trials per task on LIBERO-10 and 100 trials
    per task on RoboTwin, totaling 500 and 1,000 episodes per policy
    condition, respectively.
    All runs used training seed 1234.
    Each PPO outer step collected 128 episodes, giving 5,120 training
    episodes per task through step 40.
    The 95\% confidence intervals for paired improvements were
    [6.2, 14.0] pp on LIBERO-10 and [12.2, 18.6] pp on RoboTwin,
    estimated using 100,000 stratified paired episode bootstrap resamples.
    These intervals quantify episode-sampling uncertainty conditional on the
    trained checkpoints, not variability across training seeds.
  }
\end{table*}

%% file: tables/runtime_benchmarks.tex
\begin{table*}[t]
  \tablestyle
  \caption{Project-reported model inference frequency for baseline and
  accelerated implementations. Speedups are recomputed from the displayed
  frequencies. These values describe model execution and should not be read as
  end-to-end robot control rates. RTC-labelled rows use the Training-time RTC
  route with prefix-conditioned inference.}
  \label{tab:runtime-results}
  \small
  \setlength{\tabcolsep}{6pt}
  \begin{tabular}{@{}ccccc@{}}
    \toprule
    Device & Model path & Baseline (Hz) & Accelerated (Hz) & Speedup \\
    \midrule
    NVIDIA A100       & GR00T             & 5.96 & 32.6 & $5.47\times$ \\
    NVIDIA A100       & $\pi_{0.5}$       & 2.20 & 21.2 & $9.64\times$ \\
    \addlinespace[3pt]
    NVIDIA RTX 5090   & GR00T             & 14.7 & 42.6 & $2.90\times$ \\
    NVIDIA RTX 5090   & GR00T + RTC       & 15.0 & 47.6 & $3.17\times$ \\
    NVIDIA RTX 5090   & $\pi_{0.5}$       & 4.52 & 31.6 & $6.99\times$ \\
    \addlinespace[3pt]
    NVIDIA RTX 4090   & $\pi_{0.5}$ + RTC & 3.40 & 19.6 & $5.76\times$ \\
    \addlinespace[3pt]
    NVIDIA AGX Orin 64GB & GR00T          & 3.20 & 7.40 & $2.31\times$ \\
    NVIDIA AGX Orin 64GB & $\pi_{0.5}$    & 1.40 & 4.40 & $3.14\times$ \\
    \bottomrule
  \end{tabular}

  \par\nobreak\vspace{4pt}
  \noindent\parbox{0.97\textwidth}{\footnotesize\raggedright
  \textbf{Measurement status.} The frozen documentation reports frequencies
  but does not yet attach the raw timing traces or a complete common protocol
  covering batch size, observation shape, denoising steps, warm-up, precision,
  synchronization, power mode, and timing boundaries. The table therefore
  supports the existence of substantial path-specific speedups, not a
  hardware-normalized comparison between models.}
\end{table*}

\begin{table*}[t]
  \tablestyle
  \caption{Random-weight numerical comparison between the original and
  accelerated GR00T action heads reported by the acceleration tests.}
  \label{tab:runtime-parity}
  \small
  \setlength{\tabcolsep}{8pt}
  \begin{tabular}{@{}ccc@{}}
    \toprule
    Compared path & Cosine similarity & Maximum absolute difference \\
    \midrule
    Plain GR00T & 0.99999237 & 0.02273726 \\
    GR00T with Training-time RTC & 0.99999380 & 0.02104545 \\
    \bottomrule
  \end{tabular}

  \par\nobreak\vspace{4pt}
  \noindent\parbox{0.97\textwidth}{\footnotesize\raggedright
  Both paths use matched randomly initialized weights, inputs, and sampling
  seeds. The test checks implementation alignment under bf16 fused operations
  and CUDA Graph replay; it is not a substitute for checkpoint-level action
  parity or closed-loop task evaluation.}
\end{table*}

%% file: tables/real_robot_benchmarks.tex
\begin{table*}[t]
  \tablestyle
  \caption{Project-reported ALOHA success rates (\%) across six tasks and seven
  policy configurations.
  Overall success pools all 90 trials per policy, rather than averaging
  task-level rates. Bold denotes the highest reported value in each metric
  column, including ties.}
  \label{tab:aloha-benchmark-results}
  \small
  \setlength{\tabcolsep}{4pt}
  \begin{tabularx}{\textwidth}{@{}P{0.19\textwidth}CCCCCCP{0.12\textwidth}@{}}
    \toprule
    Policy & \multicolumn{2}{c}{Deformable-object folding}
      & \multicolumn{4}{c}{Pick-and-place} & Overall (\%) \\
    \cmidrule(lr){2-3}\cmidrule(lr){4-7}
      & Clothes & Red towel & Bear & Car & Caterpillar & Bird & \\
    \midrule
    GR00T N1.5 & 15.00 & 26.67 & 80.00 & 60.00 & 90.00 & 90.00 & 47.78 \\
    GR00T N1.7 & 5.00 & \textbf{66.67} & 90.00 & 70.00 & 90.00 & \textbf{100.00} & 62.22 \\
    $\pi_{0.5}$ & \textbf{55.00} & 50.00 & 90.00 & \textbf{100.00} & \textbf{100.00} & \textbf{100.00} & \textbf{72.22} \\
    FastWAM & 40.00 & 43.33 & \textbf{100.00} & 70.00 & 80.00 & 90.00 & 61.11 \\
    SmolVLA & 25.00 & 0.00 & 10.00 & 0.00 & 10.00 & 10.00 & 8.89 \\
    Cosmos3 & 50.00 & 23.33 & 50.00 & 60.00 & 50.00 & 40.00 & 41.11 \\
    DiT4DiT (RTC) & 25.00 & 43.33 & 90.00 & 70.00 & 70.00 & 80.00 & 54.44 \\
    \bottomrule
  \end{tabularx}

  \par\nobreak\vspace{4pt}
  \noindent\parbox{0.97\textwidth}{\footnotesize\raggedright
  Folding uses 20 clothes trials and 30 red-towel trials. Each pick-and-place
  task uses 10 trials, with a bear, a white racing car, a purple caterpillar
  toy, and a brown bird toy, respectively.
  DiT4DiT is the RTC-enabled configuration.}
\end{table*}

\begin{table*}[t]
  \tablestyle
  \caption{Project-reported Oli completion rates (\%) across four tasks.
  Box transport reports cumulative completion of stages 1--3 and full
  four-stage success separately. Bold denotes the highest reported value
  in each metric column. Dashes indicate unreported results, not failures.}
  \label{tab:oli-benchmark-results}
  \small
  \setlength{\tabcolsep}{5pt}
  \begin{tabularx}{\textwidth}{@{}P{0.16\textwidth}CCCCC@{}}
    \toprule
    Policy & \multicolumn{2}{c}{Stationary manipulation}
      & \multicolumn{3}{c}{Mobile manipulation} \\
    \cmidrule(lr){2-3}\cmidrule(lr){4-6}
      & Candy picking & Tabletop clearing
      & \multicolumn{2}{c}{Box transport} & Basket-and-toy picking \\
    \cmidrule(lr){4-5}
      & & & Stages 1--3 & All 4 stages & $\geq 2$ toys \\
    \midrule
    GR00T N1.5 & 40.83 & 43.3 & 60.0 & 6.7 & 30.0 \\
    $\pi_{0.5}$ & \textbf{62.5} & \textbf{45.0} & \textbf{61.67} & \textbf{33.33} & 28.33 \\
    FastWAM & 43.3 & 0.0 & 60.0 & 31.7 & \textbf{48.3} \\
    Cosmos3 & 15.8 & 0.0 & 56.7 & 6.7 & 11.67 \\
    DiT4DiT & 52.5 & 15.0 & -- & -- & -- \\
    \bottomrule
  \end{tabularx}

  \par\nobreak\vspace{4pt}
  \noindent\parbox{0.97\textwidth}{\footnotesize\raggedright
  $\pi_{0.5}$ box transport records 37/60 completions of stages 1--3 and
  20/60 completions of all four stages (61.67\% and 33.33\%).
  Cosmos3 basket-and-toy success is 7/60 (11.67\%). Other percentages
  retain the reported precision. FastWAM's basket-and-toy result starts
  from an intermediate state, not the full navigation sequence.}
\end{table*}

\begin{table*}[t]
  \tablestyle
  \caption{Project-reported Rebot success rates (\%) across four tasks and
  seven policy configurations. Overall success pools all 80 trials per policy.
  Bold denotes the highest reported value in each metric column.}
  \label{tab:rebot-benchmark-results}
  \small
  \setlength{\tabcolsep}{5pt}
  \begin{tabularx}{\textwidth}{@{}P{0.18\textwidth}CCCCP{0.12\textwidth}@{}}
    \toprule
    Policy & Block transfer & Box opening and car placement
      & Flower insertion & Black-block stacking & Overall (\%) \\
    \midrule
    GR00T N1.5 & 50.00 & 45.00 & 40.00 & 55.00 & 47.50 \\
    GR00T N1.7 & 75.00 & 80.00 & 55.00 & \textbf{70.00} & 70.00 \\
    $\pi_{0.5}$ & \textbf{95.00} & \textbf{85.00} & \textbf{75.00} & 55.00 & \textbf{77.50} \\
    FastWAM & 70.00 & 65.00 & 50.00 & 50.00 & 58.75 \\
    SmolVLA & 40.00 & 25.00 & 15.00 & 25.00 & 26.25 \\
    Cosmos3 & 70.00 & 70.00 & 50.00 & 40.00 & 57.50 \\
    DiT4DiT & 75.00 & 45.00 & 55.00 & 55.00 & 57.50 \\
    \bottomrule
  \end{tabularx}
\end{table*}

\begin{table*}[t]
  \tablestyle
  \caption{Project-reported dual-arm Franka success rates (\%) across five
  tasks and seven policy configurations. Overall success pools all evaluated
  trials. Bold denotes the highest reported value in each metric column.}
  \label{tab:franka-benchmark-results}
  \small
  \setlength{\tabcolsep}{5pt}
  \begin{tabularx}{\textwidth}{@{}P{0.18\textwidth}CCCCCP{0.12\textwidth}@{}}
    \toprule
    Policy & Tube handover & Bag loading & Alternating toy picking
      & Drawer loading & Block stacking & Overall (\%) \\
    \midrule
    GR00T N1.5 & 80.00 & 65.00 & 50.00 & 40.00 & 50.00 & 57.00 \\
    GR00T N1.7 & 80.00 & \textbf{95.00} & \textbf{95.00} & \textbf{90.00} & 80.00 & \textbf{88.00} \\
    $\pi_{0.5}$ & \textbf{85.00} & 75.00 & 75.00 & 75.00 & \textbf{85.00} & 79.00 \\
    FastWAM & 75.00 & 50.00 & 70.00 & 45.00 & 50.00 & 58.00 \\
    SmolVLA & 60.00 & 30.00 & 10.00 & 0.00 & 10.00 & 28.57 \\
    Cosmos3 & 80.00 & 50.00 & 60.00 & 50.00 & 45.00 & 57.00 \\
    DiT4DiT & 80.00 & 45.00 & 45.00 & 50.00 & 60.00 & 56.00 \\
    \bottomrule
  \end{tabularx}
\end{table*}

\begin{table*}[t]
  \tablestyle
  \caption{Real-robot results reported by the companion ARM
  study~\cite{arm2026}. All methods use a GR00T-N1.5 policy on an eight-stage
  AgileX ALOHA towel-folding task with a 120-second timeout. The training set
  contains 972 episodes: 809 expert and 163 DAgger episodes.
  Bold denotes the highest value in each metric column.}
  \label{tab:arm-real-robot-results}
  \small
  \setlength{\tabcolsep}{6pt}
  \begin{tabular}{@{}ccccc@{}}
    \toprule
    Method & Reward model & Success (\%) & Throughput (episodes/h) & Folding precision \\
    \midrule
    Behavior cloning & None & 62.1 & 18 & 2.2 \\
    RA-BC & SARM & 78.5 & 24 & 2.7 \\
    AW-BC & ARM & \textbf{99.4} & \textbf{32} & \textbf{3.6} \\
    \bottomrule
  \end{tabular}

  \par\nobreak\vspace{4pt}
  \noindent\parbox{0.97\textwidth}{\footnotesize\raggedright
  \textbf{Provenance.} Values are transcribed from Table~2 of the ARM paper,
  not regenerated by this report. The source defines task success and timeout
  but does not report the number of policy-evaluation trials. The result is
  evidence for one integrated data--reward--training--deployment workflow, not
  a standardized comparison across embodiments.}
\end{table*}

%% file: sections/06_conclusion.tex
\section{Conclusion and Outlook}
\label{sec:conclusion}

\subsection{Conclusion}

Embodied policy learning is no longer defined by a single modeling paradigm.
Vision--language--action models connect multimodal instructions directly to
robot actions, world--action models incorporate predicted visual dynamics and
longer temporal structure, and offline reinforcement learning or
reward-guided methods improve policies from previously collected experience.
Online RL complements these approaches by optimizing pretrained policies
through new environment interactions. Although their learning objectives
differ, these approaches depend on the same
engineering foundations: versioned data and action semantics, scalable
training, reproducible evaluation, efficient inference, and reliable
connections to simulators and physical robots. \fluxvla addresses this shared
systems problem through a configuration-driven platform that treats the full
policy lifecycle as one reconstructable workflow.

A common Parquet-oriented data layer and ordered transform pipelines separate
storage from policy-specific semantics. Registry-built models accommodate
integrated VLMs, compositional backbones and action heads, temporal or
world-model components, and reward- or advantage-weighted objectives. Runners
own distributed training, evaluation, serving, checkpoint state, and
control-loop context, while operators isolate simulator and robot
communication. Training and evaluation are connected through a clean
eval-after-train transition; resolved configurations, normalization statistics,
and complete model weights remain attached to the run; and local, remote, and
accelerated inference share the same external action-prediction contract.
Real-Time Chunking coordinates asynchronous prediction with ongoing execution,
while runner-side trajectory post-processing adapts denormalized chunks to
embodiment-specific motion requirements and active trajectories.
An optional RLinf frontend reuses native flow policies and observation--action
adapters within a distributed rollout-and-update loop, extending the platform
from fixed-data learning to interactive post-training.

The available evidence demonstrates the breadth and practical value of these
contracts while defining their present limits. Released integrations cover
autoregressive VLA policies, modular VLM/action-head architectures, pretrained
action experts, compact policies, and world--action or generative models.
Project-reported results show strong closed-loop performance on LIBERO and
working RoboCasa GR-1 paths, complemented by RoboTwin Clean and Random
results across eight integrated policies and RoboDojo progress/success
results across eight integrations on 42 tasks. The latter expose substantial
remaining gaps in full-task completion and randomized-layout generalization.
Optimized GR00T and $\pi_{0.5}$ implementations
provide substantial inference-frequency gains across datacenter, desktop, and
edge GPUs, with the available parity test showing close numerical alignment.
The six-task ALOHA benchmark provides complete results for seven VLA and
world--action policy configurations. The four-task Oli benchmark adds
stationary and mobile manipulation evidence for five policy configurations,
with partial task coverage and setup differences explicitly identified.
Four-task Rebot results further extend physical evaluation to seven policy
configurations, covering object transfer, box opening and placement, flower
insertion, and stacking.
Five-task dual-arm Franka results add bimanual handover, complementary-arm
placement, alternating grasping, drawer manipulation, and stacking across
seven policy configurations.
Reward modeling and reweighted behavior-cloning results provide an initial
example of offline, advantage-guided policy improvement within the same data
and deployment loop. These results establish \fluxvla as an engineering
foundation; they do not transfer ownership of the integrated algorithms to the
platform or imply equal statistical validation for every configuration.
The online RL integration additionally reports paired SmolVLA success-rate
improvements of 10.0 percentage points on LIBERO-10 and 15.4 percentage
points on ten RoboTwin clean tasks after 40 PPO steps
(\cref{tab:rl-post-training-results}). These task-specific results do not
establish uniform gains across all integrated policies or benchmarks.

\subsection{Outlook: Toward an integrated FluxVLA ecosystem}
\label{sec:ecosystem-outlook}

The next step is to develop \fluxvla from a unified learning framework into the
contract layer of a broader embodied-intelligence ecosystem. Rather than
incorporating data engineering, simulation, interactive collection, evaluation,
and model compilation into a single repository, these responsibilities should
remain modular and communicate through stable data, model, action, and
evaluation interfaces. This organization allows each subsystem to evolve
independently while preserving an end-to-end workflow.

\Cref{tab:flux-ecosystem} summarizes the intended division of responsibilities.
The components are currently at different stages of maturity; the table
describes their target roles rather than implying that every capability is
already production-complete.

\begin{table*}[t]
  \tablestyle
  \caption{Target organization of the FluxVLA ecosystem. Each subsystem owns a
  distinct stage while exchanging versioned artifacts through shared
  contracts.}
  \label{tab:flux-ecosystem}
  \small
  \setlength{\tabcolsep}{5pt}
  \begin{tabularx}{\linewidth}{@{}P{0.18\linewidth}CC@{}}
    \toprule
    System & Primary responsibility & Principal artifact or contract \\
    \midrule
    FluxMimir
      & Offline data ingestion, canonicalization, enrichment, validation, and
        versioned release
      & Validated Parquet/LeRobot-compatible datasets with explicit schemas,
        statistics, and provenance \\
    \addlinespace[3pt]
    FluxBisim
      & Scalable simulation, task construction, demonstration generation, and
        closed-loop rollout
      & Versioned task definitions, trajectories, simulator metadata, and
        rollout records \\
    \addlinespace[3pt]
    FluxDAgger
      & Human-in-the-loop rollout, intervention, correction, and iterative data
        collection
      & Intervention-aligned trajectories with source-policy and correction
        provenance \\
    \addlinespace[3pt]
    \fluxvla Engine
      & VLA and WAM construction, offline and online policy learning,
        distributed training, and policy inference
      & Self-contained model artifacts with preprocessing, normalization, and
        action semantics \\
    \addlinespace[3pt]
    FluxThemis
      & Simulation and physical-robot evaluation orchestration
      & Versioned evaluation protocols, trial records, metrics, and failure
        reports \\
    \addlinespace[3pt]
    FluxHermes
      & Model export, optimization, quantization, validation, and deployment
        packaging
      & Hardware-specific deployment bundles with numerical and performance
        validation \\
    \bottomrule
  \end{tabularx}
\end{table*}

\paragraph{FluxMimir: validated offline data engineering.}
Raw robot trajectories are rarely ready for direct model training. They may
differ in schema, temporal alignment, action semantics, calibration, task
annotation, and data quality. FluxMimir is designed to sit upstream of
\fluxvla and convert these heterogeneous sources into validated, versioned
dataset releases. It owns canonical conversion, semantic and temporal
validation, dataset statistics, optional visual or language annotation, and
data provenance. \fluxvla can consequently focus on model-facing transforms
and learning from a fixed dataset release rather than repeatedly implementing
source-specific conversion logic.

\paragraph{FluxBisim: scalable simulation and data generation.}
FluxBisim provides the simulation layer for constructing manipulation tasks,
generating demonstrations, and evaluating policies before physical deployment.
Its observations, actions, task definitions, and trajectories should follow the
same semantic contracts used by the learning engine and physical datasets.
This enables simulated experience to pass through FluxMimir and enter the
standard \fluxvla training pipeline. Future development can expand task
diversity, domain randomization, whole-body scenarios, and simulation-to-real
validation without introducing simulator-specific logic into individual
policies.

\paragraph{FluxDAgger: interactive policy improvement.}
FluxDAgger closes the loop between deployment failures and subsequent training.
It decouples policy inference from human takeover so that different VLA, WAM,
or reward models can share the same interactive collection workflow.
Autonomous trajectories, intervention boundaries, corrective commands, and
the identity of the generating policy can be recorded together and passed to
FluxMimir for validation and release. The resulting data can support behavior
cloning, offline reinforcement learning, reward-guided optimization, and
iterative policy refinement.

A key future direction is to preserve complete lineage across this loop. Each
correction should remain associated with the original observation, policy
prediction, human intervention, execution result, and resulting dataset
revision. This makes it possible to determine not only whether a policy
improves, but also which failures and corrections produced that improvement.

\paragraph{FluxThemis: unified evaluation orchestration.}
FluxThemis is intended to provide a dedicated evaluation layer for both
simulation and physical robots. It separates policy inference from environment
execution, rendering, scheduling, and result aggregation. This separation is
particularly useful for computationally expensive simulators, remote inference,
asynchronous checkpoint evaluation, and experiments involving shared physical
hardware.

The evaluation contract should specify task versions, initialization
conditions, trial budgets, success criteria, random seeds, intervention rules,
and failure categories. FluxThemis can then evaluate either a training
checkpoint or a deployment bundle under the same declared protocol, producing
comparable trial-level records rather than relying on manually collected
summaries.

\paragraph{FluxHermes: verified compilation and deployment.}
FluxHermes is intended to bridge the gap between a trained \fluxvla model and
an efficient hardware-specific deployment artifact. Its scope includes model
export, graph optimization, reduced-precision execution, backend compilation,
and numerical validation. The resulting bundle should preserve not only model
parameters, but also the preprocessing, normalization, embodiment, and action
semantics required to execute the policy correctly.

Performance optimization must remain subordinate to semantic fidelity.
Compiled or quantized models should therefore be validated against their source
implementations before physical deployment. Numerical agreement, latency,
memory consumption, simulation regression, and controlled robot tests together
provide stronger evidence than inference speed alone.

\subsection{A unified future lifecycle}
\label{sec:future-lifecycle}

Together, these systems form a closed improvement cycle. FluxBisim generates
simulated tasks, demonstrations, and rollout experience, while FluxDAgger
collects intervention and correction data from policy execution. FluxMimir
converts these heterogeneous sources into validated and versioned dataset
releases. \fluxvla consumes those releases to train VLA, WAM, and offline or
reward-guided policies and produces self-contained model artifacts. FluxHermes
can compile selected models into target-specific deployment bundles, and
FluxThemis evaluates checkpoints or compiled bundles in simulation and on
physical robots. The optional RL frontend also enables policies to improve
through simulator interaction, complementing dataset-based refinement;
exporting its backend checkpoints as fully self-contained deployment bundles
remains a separate integration step. Failures and interventions observed
during evaluation are returned through FluxDAgger and FluxMimir to form the
next dataset revision.